\documentclass{article}
\PassOptionsToPackage{numbers}{natbib}

\usepackage[final]{neurips_2023}

\usepackage[utf8]{inputenc} 
\usepackage[T1]{fontenc}    
\usepackage{hyperref}       
\usepackage{url}            
\usepackage{booktabs}       
\usepackage{amsfonts}       
\usepackage{nicefrac}       
\usepackage{microtype}      
\usepackage{multirow}

\usepackage{researchpack}

\usepackage{thmtools}
\declaretheorem[name=Theorem]{theorem}
\declaretheorem[name=Lemma, numberlike=theorem]{lemma}
\declaretheorem[name=Proposition, numberlike=theorem]{proposition}

\declaretheorem[name=Remark, numberlike=theorem]{remark}
\declaretheorem[name=Definition]{definition}
\declaretheorem[name=Example]{example}

\usepackage[capitalize, nameinlink]{cleveref}
\usepackage[table, dvipsnames, xcdraw]{xcolor}
\usepackage{enumitem}
\usepackage{xspace}
\usepackage{wrapfig}
\usepackage{multirow}
\usepackage{colortbl}
\usepackage{tikz}
\usetikzlibrary{positioning,shapes,arrows,calc}
 
\hypersetup{
    colorlinks=true,
    citecolor=blue,
    linkcolor=blue,
    filecolor=blue,
    urlcolor=blue,
}

\definecolor{bboxcolor}{HTML}{F626DB}

\newcommand{\changed}[1]{\textcolor{black}{#1} }

\definecolor{Gray}{gray}{0.9}

\newcommand{\MNISTShortcut}{{\tt MNIST-EvenOdd}\xspace}
\newcommand{\MNISTAdd}{{\tt MNIST-Addition}\xspace}
\newcommand{\MNISTMul}{{\tt MNIST-Multiplication}\xspace}
\newcommand{\MNISTOp}{{\tt MNIST-AddMul}\xspace}
\newcommand{\XOR}{{\tt XOR}\xspace}
\newcommand{\BOIA}{{\tt BDD-OIA}\xspace}

\newcommand{\MSR}{\textsc{r}\xspace}
\newcommand{\MSC}{\textsc{c}\xspace}
\newcommand{\MSH}{\textsc{h}\xspace}
\newcommand{\MTL}{\textsc{mtl}\xspace}
\newcommand{\DIS}{\textsc{dis}\xspace}

\newcommand{\SL}{\ensuremath{\mathsf{SL}}\xspace}

\newcommand{\MZero}{\includegraphics[width=1.85ex]{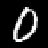}\xspace}
\newcommand{\MOne}{\includegraphics[width=1.85ex]{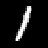}\xspace}
\newcommand{\MTwo}{\includegraphics[width=1.85ex]{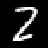}\xspace}
\newcommand{\MThree}{\includegraphics[width=1.85ex]{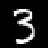}\xspace}
\newcommand{\MFour}{\includegraphics[width=1.85ex]{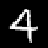}\xspace}
\newcommand{\MFive}{\includegraphics[width=1.85ex]{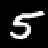}\xspace}
\newcommand{\MSix}{\includegraphics[width=1.85ex]{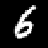}\xspace}
\newcommand{\MSeven}{\includegraphics[width=1.85ex]{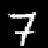}\xspace}
\newcommand{\MEight}{\includegraphics[width=1.85ex]{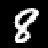}\xspace}
\newcommand{\MNine}{\includegraphics[width=1.85ex]{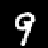}\xspace}

\newcommand{\dataset}{\ensuremath{\calD}\xspace}

\newcommand{\BK}{\ensuremath{\mathsf{K}}\xspace}
\newcommand{\KL}{\ensuremath{\mathsf{KL}}\xspace}
\newcommand{\Ent}{\ensuremath{\mathsf{H}}\xspace}
\newcommand{\MI}{\ensuremath{\mathsf{I}}\xspace}

\newcommand{\shortsection}[1]{\vspace{-3pt}\section{#1}\vspace{-5pt}}

\title{Not All Neuro-Symbolic Concepts Are Created Equal:\\ Analysis and Mitigation of Reasoning Shortcuts}

\author{%
    Emanuele Marconato \\
    DISI and DI \\
    University of Trento and University of Pisa \\
    Trento, Italy \\
    \texttt{emanuele.marconato@unitn.it} \\
    \And
    Stefano Teso \\
    CIMeC and DISI \\
    University of Trento \\
    Trento, Italy \\
    \texttt{stefano.teso@unitn.it} \\
    \And
    Antonio Vergari \\
    School of Informatics \\
    University of Edinburgh \\
    Edinburgh, UK \\
    \texttt{avergari@exseed.ed.ac.uk} \\
    \And
    Andrea Passerini \\
    DISI \\
    University of Trento \\
    Trento, Italy \\
    \texttt{andrea.passerini@unitn.it} \\
}

\begin{document}
\maketitle

\begin{abstract}
    Neuro-Symbolic (NeSy) predictive models hold the promise of improved compliance with given constraints, systematic generalization, and interpretability, as they allow to infer labels that are consistent with some prior knowledge by reasoning over high-level concepts extracted from sub-symbolic inputs.
    It was recently shown that NeSy predictors are affected by \textit{reasoning shortcuts}: they can attain high accuracy but by leveraging concepts with \textit{unintended semantics}, thus coming short of their promised advantages.
    Yet, a systematic characterization of reasoning shortcuts and of potential mitigation strategies is missing.
    This work fills this gap by characterizing them as unintended optima of the learning objective and identifying four key conditions behind their occurrence.
    Based on this, we derive several natural mitigation strategies, and analyze their efficacy both theoretically and empirically.
    Our analysis shows reasoning shortcuts are difficult to deal with, casting doubts on the trustworthiness and interpretability of existing NeSy solutions.
\end{abstract}

\section{Introduction}

Neuro-Symbolic (NeSy) AI  aims at improving the \textit{robustness} and \textit{trustworthiness} of neural networks by integrating them with reasoning capabilities and prior knowledge \citep{de2021statistical, garcez2022neural}.
We focus on \textit{NeSy predictors} \citep{giunchiglia2022deep, dash2022review}, neural structured-output classifiers that infer one or more labels by reasoning over high-level \textit{concepts} extracted from sub-symbolic inputs, like images or text \citep{diligenti2017semantic, donadello2017logic, manhaeve2018deepproblog, xu2018semantic, giunchiglia2020coherent, ahmed2022semantic, li2023learning}.
They leverage reasoning techniques to encourage -- or even \textit{guarantee} -- that their predictions comply with domain-specific regulations and inference rules.
As such, they hold the promise of improved \textit{systematic generalization}, \textit{modularity}, and \textit{interpretability}, in that learned concepts can be readily reused in different NeSy tasks, as done for verification \citep{xie2022neuro}, and for explaining the model's inference process to stakeholders \citep{rudin2019stop, chen2019looks, chen2020concept}.
On paper, this makes NeSy predictors ideal for high-stakes applications that require both transparency and fine-grained control over the model's (in- and out-of-distribution) behavior, such as medical diagnosis \citep{degrave2021ai}, robotics \citep{maiettini2019weakly} and self-driving cars \citep{badue2021self}.

\begin{figure}[!t]
    \centering
    \includegraphics[width=0.95\textwidth]{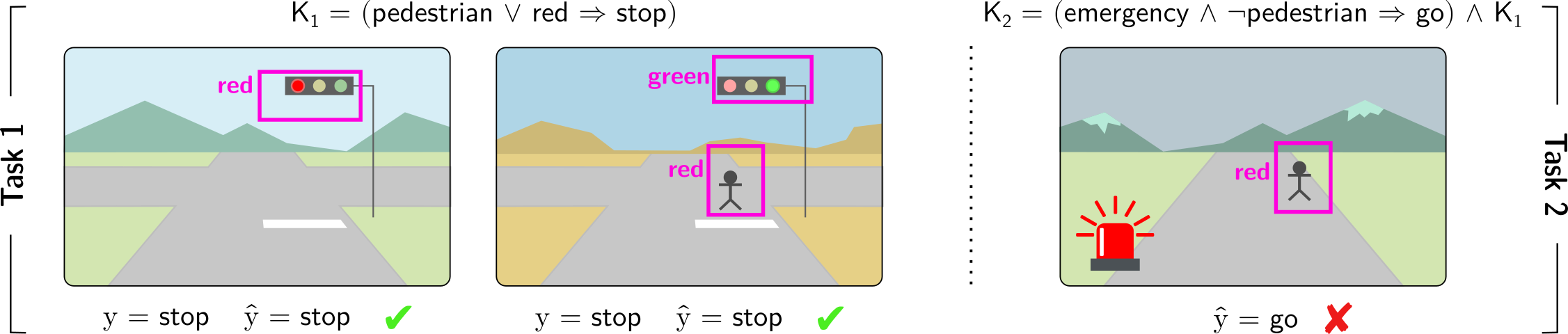}
    \vspace{-0.5em}
    \caption{
    \textbf{Reasoning shortcuts undermine trustworthiness}.\scalebox{0.01}{chiodo fisso}  An autonomous vehicle has to decide whether to $Y = {\tt stop}$ or $Y = {\tt go}$ based on three binary concepts extracted from an image $\vx$, namely $C_1 = {\tt red}$ light, $C_2 = {\tt green}$ light and $C_3 =$ presence of {\tt pedestrian}s (shown in \textbf{\textcolor{bboxcolor}{pink}}).
    \textbf{Left}: In Task 1, the prior knowledge $\BK = ({\tt pedestrian} \lor {\tt red} \Rightarrow {\tt stop})$ instructs the vehicle to stop whenever the light is red or there are pedestrians on the road.
    The model can perfectly classify an (even exhaustive) training set by acquiring a \textit{reasoning shortcut that classifies pedestrians as red lights}.
    \textbf{Right}:  The learned concepts are then reused to guide an autonomous ambulance with the additional rule that in emergency situations red lights can be ignored, with potentially dire consequences.\includegraphics[scale=0.001]{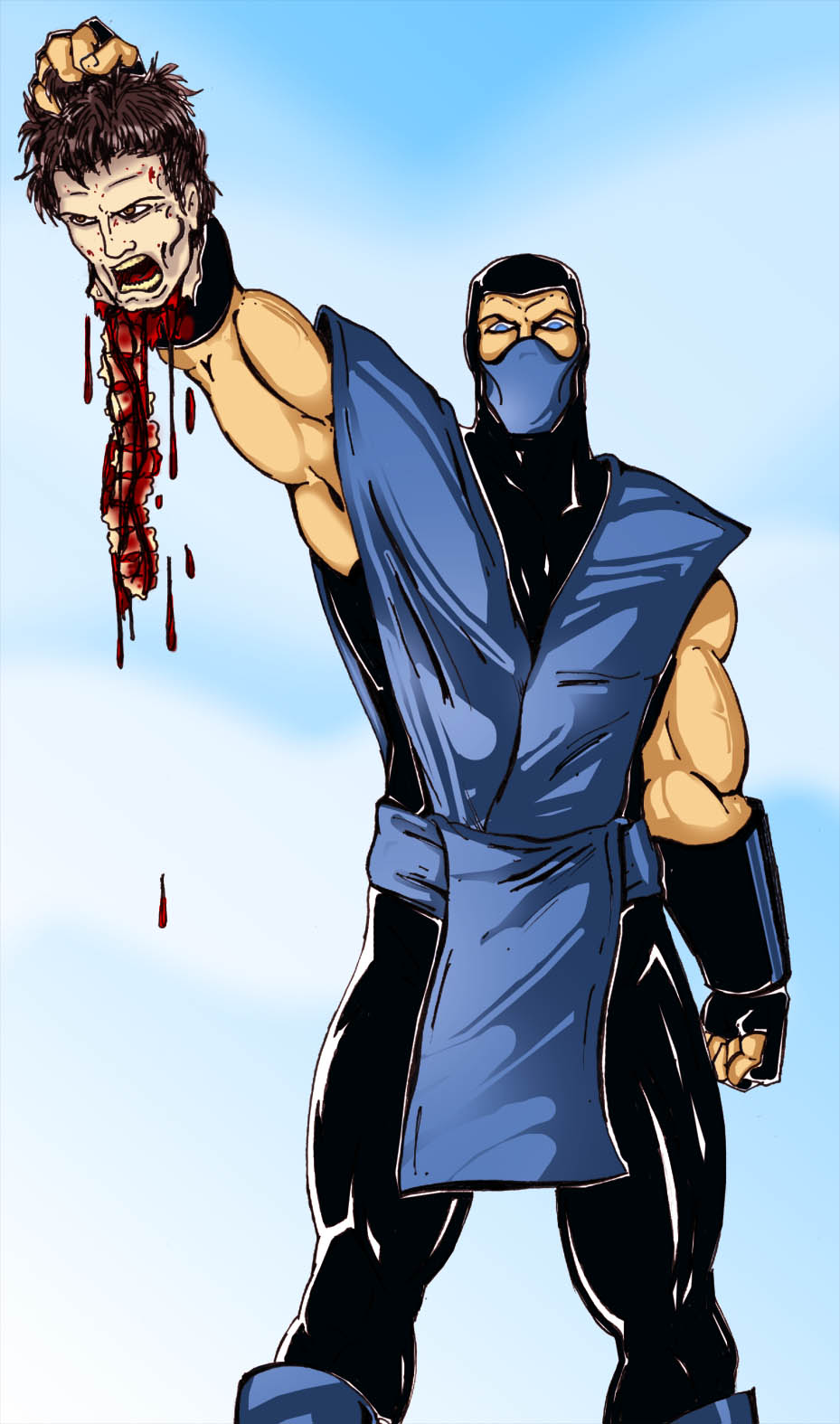}
    Our work identifies the causes of RSs (\cref{sec:properties}) and  several mitigation strategies (\cref{sec:mitigation-strategies}).
    }
    \label{fig:second-page}
\end{figure}

Much of the promise of these models relies on learned concepts being \textit{high quality}.
The general consensus is that the prior knowledge constrains learned concepts to behave as expected \citep{fredrikson2023learning} and issues with them are often tackled heuristically \citep{manhaeve2021neural}.
It was recently shown that, however, NeSy predictors can \textit{attain high accuracy by leveraging concepts with unintended semantics}~\citep{marconato2023neuro}.  Following \citet{marconato2023neuro}, we refer to these as \textit{reasoning shortcuts} (RSs).
RSs are problematic, as concepts encoding unintended semantics compromise generalization across NeSy tasks, as shown in \cref{fig:second-page}, as well as interpretability and verification of NeSy systems~\citep{xie2022neuro}.
Moreover, the only known mitigation strategies are based on heuristics \citep{manhaeve2021neural, li2023learning}.

The issue is that RSs -- and their root causes -- are not well understood, making it difficult to design effective remedies.
In this paper, we fill this gap.
We introduce a formal definition of RSs and theoretically characterize their properties, highlighting how they are a general phenomenon affecting a variety of state-of-the-art NeSy predictors.
Specifically, our results show that RSs can be shared across many NeSy  architectures, and provide a way of \textit{counting} them for any given learning problem. 
They also show that RSs depend on \textit{four key factors}, namely the structure of the prior knowledge and of the data, the learning objective, and the architecture of the neural concept extractor.
This enables us to identify several supervised and unsupervised mitigation strategies, which we systematically analyze both theoretically and empirically.
Finally, we experimentally validate our findings by testing a number of representative NeSy predictors and mitigation strategies on four NeSy data sets.

\textbf{Contributions.}  Summarizing, we:
(\textit{i}) Formalize RSs and identify four key root causes.
(\textit{ii}) Show that RSs are a general issue impacting a variety of NeSy predictors.
(\textit{iii}) Identify a number of mitigation strategies and analyze their effectiveness, or lack thereof.
(\textit{iv}) Empirically show that RSs arise even when the data set is large and unbiased, and evaluate the efficacy of different mitigation strategies, highlighting the limits of unsupervised remedies and the lack of a widely applicable recipe.

\shortsection{The Family of Neuro-Symbolic Predictors}
\label{sec:nesy-predictors}

\textbf{Notation.}  Throughout, we indicate scalar constants $x$ in lower-case, random variables $X$ in upper case, and ordered sets of constants $\vx$ and random variables $\vX$ in bold typeface.
Also, $\vx_{i:j}$ denotes the subset $\{x_i, \ldots, x_j\}$, $[n]$ the set $\{1, \ldots, n\}$, and $\vx \models \BK$ indicates that $\vx$ satisfies a logical formula $\BK$.

\textbf{NeSy predictors.}  A \textit{NeSy predictor} is a model that infers $n$ \textit{labels} $\vY$ (taking values in $\calY$) by reasoning over a set of $k$ discrete \textit{concepts} $\vC$ (taking values in $\calC$) extracted from a {sub-symbolic} continous \textit{input} $\vX$ (taking values in $\calX$).
Reasoning can be implemented in different ways, but overall its role is to encourage the model's predictions to comply with given \textit{prior knowledge} $\BK$, in the sense that predictions $\vy$ that do not satisfy the knowledge are generally avoided.
The prior knowledge $\BK$ is assumed to be provided upfront and correct, as formalized in \cref{sec:reasoning-shortcuts}.
Normally, only supervision on the labels $\vY$ is available for training, with the concepts $\vC$ treated as latent variables.

\begin{example}
\label{ex:mnist-addition}
In \MNISTAdd~\citep{manhaeve2018deepproblog}, given a pair of MNIST images \citep{lecun1998mnist}, say $\vx = ( \MTwo, \MSix )$, the model has to infer the concepts  $\vC = (C_1, C_2)$ encoding the digit classes, to predict their sum $Y$, in this case $8$.  Reasoning drives the model towards complying with constraint $\BK = (Y = C_1 + C_2)$.
\end{example}

The concepts are modeled by a conditional distribution $p_\theta(\vC \mid \vX)$ parameterized by $\theta \in \Theta$, typically implemented with a neural network.
The predicted concepts can be viewed as ``soft'' or ``neural'' predicates with a truth value ranging in $[0,1]$.
As for the reasoning step, the most popular strategies involve \textit{penalizing} the model for producing concepts and/or labels inconsistent with the knowledge at training time~\citep{xu2018semantic, fischer2019dl2, ahmed2022neuro} or introducing a \textit{reasoning layer} that infers labels from the predicted concepts and also operates at inference time~\citep{manhaeve2018deepproblog, giunchiglia2020coherent, hoernle2022multiplexnet, ahmed2022semantic}.
In either case, end-to-end training requires to differentiate through the knowledge.  Mainstream options include softening the knowledge using fuzzy logic~\citep{diligenti2012bridging, donadello2017logic, pryor2022neupsl, li2023learning} and casting reasoning in terms of probabilistic logics~\citep{de2015probabilistic, manhaeve2018deepproblog, ahmed2022semantic}.

To investigate the scope and impact of RSs, we consider three representative NeSy predictors.
The first one is \textbf{DeepProbLog} (DPL) \citep{manhaeve2018deepproblog}, which implements a sound probabilistic-logic reasoning layer on top of the neural predicates.
DPL is a discriminative predictor of the form~\citep{marconato2023neuro}:
\[
    \textstyle
    p_\theta(\vy \mid \vx; \BK)
        = \sum_\vc u_\BK(\vy \mid \vc) \cdot p_\theta(\vc \mid \vx)
    \label{eq:dpl-likelihood}
\]
where the concept distribution is fully factorized and the label distribution is \textit{uniform}\footnote{In practice, the distribution of labels given concepts needs not be uniform \citep{ahmed2022semantic}.} over all label-concept combinations compatible with the knowledge, that is, $u_\BK(\vy \mid \vc) = \Ind{\vc \models \BK[\vY/\vy]} / Z(\vc; \BK)$
where the indicator $\Ind{ \vc \models \BK[\vY/\vy] }$ \textit{guarantees} all labels inconsistent with $\BK$ have zero probability, and $Z(\vc; \BK)= {\textstyle \sum_\vy} \Ind{\vc \models \BK[\vY/\vy]}$ is a normalizing constant.
Inference amounts to computing a most likely label $\argmax_{\vy} \ p_\theta(\vy \mid \vx ; \BK)$, while learning is carried out via maximum (log-)likelihood estimation, that is given a training set $\dataset = \{ (\vx, \vy) \}$, maximizing
\[
    \textstyle
    \calL(p_\theta, \dataset, \BK) := \frac{1}{|\dataset|}
            \sum_{ (\vx, \vy) \in \dataset } \ \log {p_\theta (\vy \mid \vx; \BK)}.
    \label{eq:dpl-learning-objective}
\]
In general, it is intractable to evaluate \cref{eq:dpl-likelihood} and solve inference exactly.  DPL leverages knowledge compilation~\citep{darwiche2002knowledge, vergari2021compositional} to make both steps practical.
%

Our analysis on DPL can be carried over to other NeSy predictors implementing analogous reasoning layers~\citep{manhaeve2021approximate, huang2021scallop, winters2022deepstochlog, ahmed2022semantic, van2022anesi}.
Furthermore, we show that certain RSs affect also alternative NeSy approaches
such as the \textbf{Semantic Loss} (SL) \citep{xu2018semantic} and \textbf{Logic Tensor Networks} (LTNs) \citep{donadello2017logic}, two state-of-the-art penalty-based approaches.
Both reward a neural network for predicting labels $\vy$ consistent with the knowledge, but SL measures consistency in probabilistic terms, while LTNs use fuzzy logic to measure a fuzzy degree of knowledge satisfaction.
See \cref{sec:other-approaches} for a full description.

\section{Reasoning Shortcuts as Unintended Optima}
\label{sec:reasoning-shortcuts}

It was recently shown that NeSy predictors are vulnerable to \textit{reasoning shortcuts} (RSs), whereby the model attains high accuracy by leveraging concepts with \textit{unintended semantics}~\citep{marconato2023neuro}.

\begin{example}
\label{ex:mnist-addition-with-shortcuts}
To build intuition, consider \MNISTAdd and assume the model is trained on examples of only two sums:  $\MZero + \MOne = 1$ and $\MZero + \MTwo = 2$.  Note that there exist \textit{two distinct maps from images to concepts that perfectly classify such examples}:  one is the intended solution $(\MZero \mapsto 0, \MOne \mapsto 1, \MTwo \mapsto 2)$, while the other is $(\MZero \mapsto 1, \MOne \mapsto 0, \MTwo \mapsto 1)$.  The latter is unintended.
\end{example}

\begin{wrapfigure}{r}{0.25\linewidth}
    \centering
    \vspace{-10pt}
    \scalebox{.85}{
    \begin{tikzpicture}[
    scale=1.15,
    transform shape,
    node distance=.35cm and .35cm,
    minimum width={width("G")+15pt},
    minimum height={width("G")+15pt},
    mynode/.style={draw,ellipse,align=center}
]
    \node[mynode] (S) {$\vS$};
    \node[mynode, right=of S] (G) {$\vG$};
    \node[mynode, below=of S, fill=black!13!white] (X) {$\vX$};
    \node[mynode, below=of G, fill=black!13!white] (Y) {$\vY$};
    \node[mynode, below=of Y, blue] (C) {$\vC$};
        
    \path
        (G) edge[-latex] (X)
        (S) edge[-latex] (X)
        (G) edge[-latex] (Y);

    \path [blue]
        (X) edge[-latex] (C)
        (C) edge[-latex] (Y);

\end{tikzpicture}
    }
    \caption{The ground-truth data generation process (in \textbf{black}) and a NeSy predictor (in \textbf{\textcolor{blue}{blue}}).}
    \label{fig:generative-process}
    \vspace{-5pt}
\end{wrapfigure}
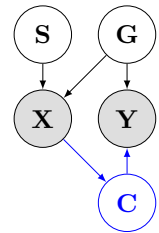

\textbf{The ground-truth data generating process.}  In order to properly define what a RS is, we have to first define what \textit{non}-shortcut solutions are.
In line with work on causal representation learning \citep{scholkopf2021toward, suter2019robustly, vonkugelgen2021self}, we do so by specifying the ground-truth data generation process $p^*(\vX, \vY ; \BK)$.
Specifically, we assume it takes the form illustrated in \cref{fig:generative-process}.
In short, we assume there exist $k$ unobserved ground-truth concepts $\vG$ (\eg in \MNISTAdd these are the digits $0$ to $9$) that determine \textit{both} the observations $\vX$ (the MNIST images) and the labels $\vY$ (the sum).
We also allow for extra stylistic factors $\vS$, independent from $\vG$, that do influence the observed data (\eg calligraphic style) but \textit{not} the labels.
The ground-truth concepts are discrete and range in $\calG = [m_1] \times \cdots \times [m_k]$, whereas the style $\vs \in \bbR^q$ is continuous.
The training and test examples $(\vx, \vy)$ are then obtained by first sampling $\vg$ and $\vs$ and then $\vx \sim p^*(\vX \mid \vg, \vs)$ and $\vy \sim p^*(\vY \mid \vg; \BK)$.
Later on, we will use $\mathsf{supp}(\vG)$ to denote the support of $p^*(\vG)$.
We also assume that the ground-truth process is consistent with the prior knowledge $\BK$, in the sense that invalid examples are never generated: $p^*(\vy \mid \vg; \BK) = 0$ for all $(\vg, \vy)$ that violate $\BK$.

\textbf{What is a reasoning shortcut?}  By definition, a NeSy predictor $p_\theta(\vY \mid \vX; \BK)$ -- shown in \textbf{\textcolor{blue}{blue}} in \cref{fig:generative-process} -- acquires concepts with the correct semantics if it recovers the ground-truth concepts, \ie
%
\[
    \textstyle
    p_\theta(\vC \mid \vx) \equiv p^*(\vG \mid \vx) \quad \forall \ \vx \in \calX
    \label{eq:concept-identifiability}
\]
A model satisfying \cref{eq:concept-identifiability} easily generalizes to other NeSy prediction tasks that make use of the same ground-truth concepts $\vG$, as $p_\theta(\vC \mid \vX)$ can be reused for solving the new task~\citep{marconato2023neuro, quinonero2008dataset}.
It is also interpretable, in the sense that as long as stakeholders understand the factors $\vG$, they can also interpret concept-based explanations of the inference process that rely on $\vC$~\citep{marconato2022glancenets}.
Naturally, there may exist many concept distributions that \textit{violate} \cref{eq:concept-identifiability} and that as such do \textit{not} capture the correct semantics.
However, all those that achieve sub-par log-likelihoods can be ideally avoided simply by improving learning or supplying more examples.
We define RSs as those concept distributions for which these strategies are not enough.

\begin{definition}
\label{def:reasoning-shortcut}
A reasoning shortcut is a distribution $p_\theta(\vC \mid \vX)$ that achieves maximal log-likelihood on the training set but does not match the ground-truth concept distribution,
\[
    \textstyle
    \calL(p_\theta, \dataset, \BK) = \max_{\theta' \in \Theta} \ \calL(p_{\theta'}, \dataset, \BK)
    \qquad \land \qquad
    p_\theta(\vC \mid \vX) \not\equiv p^*(\vG \mid \vX)
    \label{eq:reasoning-shortcut}
\]
\end{definition}

This makes RSs difficult to improve by regular means and also hard to spot based on predictions alone.
Yet, since the concepts do not recover the correct semantics, RSs can compromise systematic generalization and interpretability.
For instance, the shortcut concepts learned in \cref{ex:mnist-addition-with-shortcuts} would fail completely at \MNISTAdd tasks involving digits other than $0$, $1$, and $2$ and also at other arithmetic tasks, like multiplication, involving the same digits, as we show experimentally in \cref{sec:experiments}.
An examples of RSs in a high-stakes scenario is shown in \cref{fig:second-page}.

\section{Properties of Reasoning Shortcuts}
\label{sec:properties}

\textbf{Counting deterministic RSs.}  We begin by assessing how many RSs exist in an idealized setting in which the ground-truth generative process is simple, we have access to the true risk, and the concepts $\vC$ have been specified correctly, that is, $\calC = \calG$, and then proceed to work out what this entails.
Specifically, we work with these assumptions:
\begin{itemize}[leftmargin=2em]

    \item[\textbf{A1}.] The distribution $p^*(\vX \mid \vG, \vS)$ is induced by a map $f: (\vg, \vs) \mapsto \vx$, \ie $p^*(\vX \mid \vG, \vS) = \delta\{\vX = f(\vg, \vs)\}$, where  $f$ is \textit{invertible}, and \textit{smooth} over $\vs$.

    \item[\textbf{A2}.] The distribution $p^*(\vY \mid \vG; \BK)$ is induced by a map $\beta_\BK: \vg \mapsto \vy$.  This is what happens in \MNISTAdd, as there exists a unique value $y$ that is the sum of any two digits $(g_1, g_2)$.

\end{itemize}
Our analysis builds on a link between the NeSy predictor $p_\theta(\vY \mid \vX; \BK)$ and the concept extractor $p_\theta(\vC \mid \vX)$, which depend on $\vX$, and their analogues $p_\theta(\vY \mid \vG; \BK)$ and $p_\theta(\vC \mid \vG)$ that depend directly on the ground-truth concepts $\vG$.
This link is formalized by the following lemma:

\begin{lemma}
    \label{lemma:abstraction-from-lh}
    It holds that:
    (\textit{i})  The true risk of $p_\theta$ can be upper bounded as follows:
    \begin{equation}
    \resizebox{.94\hsize}{!}{
    $\textstyle
        \bbE_{(\vx, \vy) \sim p^*(\vX , \vY ; \BK)} [
            \log p_\theta(\vy \mid \vx; \BK)
        ]
        \leq
        \bbE_{\vg \sim p(\vG)} \big(
            - \KL [ p^*(\vY \mid \vg; \BK) \ \| \ p_\theta(\vY \mid \vg; \BK) ] - \Ent [ p^*(\vY \mid \vg; \BK) ]
        \big)$
    }
    \label{eq:dpl-upper-bound}
    \end{equation}
    where \KL is the Kullback-Leibler divergence and \Ent is the Shannon entropy.  Moreover, under \textbf{A1} and \textbf{A2}, $p_\theta(\vY \mid \vX; \BK)$ is an optimum of the LHS of \cref{eq:dpl-upper-bound} if and only if $p_\theta(\vY \mid \vG; \BK)$ is an optimum of the RHS.
    (\textit{ii}) Under \textbf{A1}, there exists a bijection between the deterministic concept distributions $p_\theta(\vC \mid \vX)$ \changed{that are constant over the support of $p(\vX \mid \vg)$, for each $\vg \in \mathrm{supp}(\vG)$,} and the deterministic distributions of the form $p_\theta(\vC \mid \vG)$.
\end{lemma}

All proofs can be found in \cref{sec:proofs}.
\cref{lemma:abstraction-from-lh} implies that the deterministic concept distributions $p_\theta(\vC \mid \vX)$ of NeSy predictors $p_\theta(\vY \mid \vX; \BK)$ that maximize the LHS of \cref{eq:dpl-upper-bound}, including those that are RSs, correspond one-to-one to the deterministic distributions $p_\theta(\vC \mid \vG)$ yielding label distributions $p_\theta(\vY \mid \vG; \BK)$ that maximize the RHS of \cref{eq:dpl-upper-bound}.
Hence, we can count the number of \textit{deterministic} RSs by counting the deterministic distributions $p_\theta(\vC \mid \vG)$:

\begin{theorem}
    \label{thm:mc-det-opts}
    Let $\calA$ be the set of mappings $\alpha: \vg \mapsto \vc$ induced by all possible deterministic distributions $p_\theta(\vC \mid \vG)$, \ie each $p_\theta(\vC \mid \vG) = \Ind{\vC = \alpha(\vG)}$ for exactly one $\alpha \in \calA$.
    Under \textbf{A1} and \textbf{A2}, the number of deterministic optima $p_\theta(\vC \mid \vG)$ of \cref{eq:dpl-upper-bound} is:
    \[ 
        \textstyle
        \sum_{\alpha \in \calA} \Ind{
            \bigwedge_{\vg \in \mathsf{supp}(\vG)}
                (\beta_\BK \circ \alpha)(\vg) = \beta_\BK(\vg)
        }
        \label{eq:model-count}
    \]
\end{theorem}

Intuitively, this sum counts the deterministic concept distributions $p_\theta(\vC \mid \vX)$ -- embodied here by the maps $\alpha$ -- that output concepts predicting a \textit{correct} label for each example in the training set.
The ground-truth distribution $p^*(\vG \mid \vX)$ is one such distribution, so the count is always at least one, but there may be more, and all of these are RSs.  \cref{eq:model-count} gives us their exact number.
This formalizes the intuition of \citet{marconato2023neuro} that, as long as the prior knowledge $\BK$ admits the correct label $\vy$ to be inferred from more than one concept vector $\vc$, there is room for RSs.
So far, we have assumed \cref{eq:dpl-learning-objective} is computed as in DPL.  However, RSs are chiefly a property of the \textit{prior knowledge}, and as such also affect NeSy predictors employing different reasoning procedures or different relaxations of the knowledge.  We show this formally in \cref{sec:other-approaches}.
Deterministic RS are also important because -- in certain cases -- they define a basis for \textit{all} reasoning shortcuts:

\begin{proposition}
    \label{prop:structure-of-nondet-ops}
    For probabilistic logic approaches (including DPL and SL):
    (\textit{i}) All convex combinations of two or more deterministic optima $p_\theta(\vC \mid \vX)$ of the likelihood are also (non-deterministic) optima. \changed{However, not all convex combinations can be expressed in DPL and SL.}
    (\textit{ii}) Under \textbf{A1} and \textbf{A2}, all optima of the likelihood can be expressed as a convex combination of deterministic optima.
    (\textit{iii}) If \textbf{A2} does not hold, there may exist non-deterministic optima that are not convex combinations of deterministic ones. These may be the only optima.
\end{proposition}

Combining \cref{prop:structure-of-nondet-ops} (\textit{i}) with \cref{thm:mc-det-opts} gives us a lower bound for the number of \textit{non-deterministic} RSs, in the sense that if there are at least two deterministic RS, then there exist infinitely many non-deterministic ones.
An important consequence is that, if we can somehow control what deterministic RSs affect the model, then we may be able to implicitly lower the number of \textit{non-deterministic} RSs as well.
However, \cref{prop:structure-of-nondet-ops} implies that there may exist \textit{non-deterministic} RSs that are unrelated to the deterministic ones and that as such cannot be controlled this way.

\section{Analysis of Mitigation Strategies}
\label{sec:mitigation-strategies}

The key factors underlying the occurrence of deterministic RSs appear explicitly in \cref{eq:model-count}.  These are:
(\textit{i}) the knowledge $\BK$,
(\textit{ii}) the structure of $\mathrm{supp}(\vG)$,
(\textit{iii}) the objective function $\calL$ used for training (via \cref{lemma:abstraction-from-lh}), and
(\textit{iv}) the architecture of the concept extractor $p_\theta(\vC \mid \vX)$, embodied in the Theorem by $p_\theta(\vC \mid \vG)$.
This gives us a starting point for identifying possible mitigation strategies and analyzing their impact on the number of \textit{deterministic} RSs.
Our main results are summarized in \cref{tab:mitigation-strategies}.

\begin{table}[!t]
    \centering
    \footnotesize
    \caption{\textbf{Impact of different mitigation strategies on the number of deterministic optima}:  \MSR is reconstruction, \MSC supervision on $\vC$, \MTL multi-task learning, and \DIS disentanglement.
    All strategies reduce the number of $\alpha$'s in \cref{eq:model-count}, sometimes substantially, but require different amounts of effort to be put in place.
    Actual counts for our data sets are reported in \cref{sec:datasets}.
    }
    \label{tab:mitigation-strategies}
    \scalebox{.9}{
    \begin{tabular}{lllcl}
        \toprule
        {\sc Mitigation}
            & {\sc Requires}
            & {\sc Constraint on $\alpha$}
            & {\sc Assumptions}
            & {\sc Result}
        \\
        \midrule
        None
            & --
            & $  {\bigwedge_{\vg \in \mathsf{supp}(\vG)} \big( (\beta_\BK \circ \alpha)(\vg) = \beta_\BK(\vg) \big) } $
            & \textbf{A1}, \textbf{A2}
            & \cref{thm:mc-det-opts}
        \\
        \MTL
            & Tasks
            & $ {\bigwedge_{\vg \in \mathsf{supp}(\vG)} \bigwedge_{t \in [T]} \big( (\beta_{\BK^{(t)}} \circ \alpha)(\vg) = \beta_{\BK^{(t)}}(\vg) \big) } $
            & \textbf{A1}, \textbf{A2}
            & \cref{prop:multitask}
        \\
        \MSC
            & Sup. on $\vC$
            &  $ {\bigwedge_{\vg \in \calS \subseteq \mathsf{supp}(\vG)} \bigwedge_{i \in I}  \big( \alpha_i(\vg) = g_i  \big) } $
            & \textbf{A1}
            & \cref{prop:concept-supervision}
        \\
        \MSR
            & --
            & $ {\bigwedge_{\vg, \vg' \in \mathsf{supp}(\vG) : \vg \ne \vg'} \big( \alpha(\vg) \neq  \alpha(\vg') \big) } $
            & \textbf{A1}, \textbf{A3}
            & \cref{prop:recon-loss}
        \\
        \bottomrule
    \end{tabular}
    }
\end{table}

\subsection{Knowledge-based Mitigation}

The \textit{prior knowledge} \BK is the main factor behind RSs and also a prime target for mitigation.
The most direct way of eliminating unintended concepts is to edit $\BK$ directly, for instance by eliciting additional constraints from a domain expert.
However, depending on the application, this may not be feasible:  experts may not be available, or it may be impossible to constrain $\BK$ without also eliminating concepts with the intended semantics.

A more practical alternative is to employ \textit{Multi-Task Learning} (\MTL).
The idea is to train a NeSy predictor over $T$ tasks sharing the same ground-truth concepts $\vG$ but differing prior knowledge $\BK^{(t)}$, for $t \in [T]$.
\textit{E.g.}, one could learn a model to predict both the sum and product of MNIST digits, as in our experiments (\cref{sec:experiments}).
Intuitively, by constraining the concepts to work well across tasks, \MTL leaves less room for unintended semantics.
The following result confirms this intuition:

\begin{proposition}
    \label{prop:multitask}
    Consider $T$ NeSy prediction tasks with knowledge $\BK^{(t)}$, for $t \in [T]$ and data sets $\dataset^{(t)}$, all sharing the same $p^*(\vG)$.
    Under \textbf{A1} and \textbf{A2}, any deterministic optimum $p_\theta(\vC \mid \vG)$ of the \MTL loss (\ie the average of per-task losses) is a deterministic optimum of a single task with prior knowledge $\bigwedge_t \BK^{(t)}$.
    The number of deterministic optima amounts to:
    \[
        \textstyle
        \sum_{\alpha \in \calA} \Ind{\bigwedge_{\vg \in \mathsf{supp}(\vG)} \bigwedge_{t =1}^T  \big( (\beta_{\BK^{(t)}} \circ \alpha)(\vg) = \beta_{\BK^{(t)}}(\vg) \big) }
        \label{eq:mc-multitask}
    \]
\end{proposition}

This means that, essentially, \MTL behaves like a logical conjunction:  any concept extractor $p_\theta(\vC \mid \vG)$ incompatible with the knowledge of \textit{any} task $t$ is not optimal.
This strategy can be very effective, and indeed it performs very well in our experiments, but it necessitates gathering or designing a \textit{set} of correlated learning tasks, which may be impractical in some situations.

\subsection{Data-based Mitigation}

Another key factor is the support of $p^*(\vG)$:  if the support is not full, the conjunction in \cref{eq:model-count} becomes looser, and the number of $\alpha$'s satisfying it increases.
This is what happens in \MNISTShortcut (\cref{ex:mnist-addition-with-shortcuts}):  here, RSs arise precisely because the training set only includes a \textit{subset} of combinations of digits, leaving ample room for acquiring unintended concepts.

We stress, however, that RSs can also occur if the data set is \textit{exhaustive}, as in the next example.

\begin{example}[\XOR task]
\label{ex:xor}
Consider a task with three binary ground-truth concepts $\vG = (G_1, G_2, G_3)$ in which the label $Y$ is the parity of these bits, that is $\BK = (Y = G_1 \oplus G_2 \oplus G_3)$.  Each label $Y \in \{0, 1\}$ can be inferred from four possible concept vectors $\vg$, meaning that knowing $\vy$ is not sufficient to identify the $\vg$ it was generated from.  In this case, it is impossible to pin down the ground-truth distribution $p^*(\vC \mid \vG; \BK)$ even if all possible combinations of inputs $\vx$ are observed.
\end{example}

One way of avoiding RSs is to explicitly guide the model towards satisfying the condition in \cref{eq:concept-identifiability} by supplying \textit{supervision} for a subset of concepts $\vC_I \subseteq \vC$, with $I \subseteq [k]$, and then augmenting the log-likelihood with a cross-entropy loss over the concepts of the form
$
    \textstyle
    \sum_{i \in I} \log p_\theta(C_i = g_i \mid \vx)
$.
Here, training examples $(\vx, \vg_I, \vy)$ come with annotations for the concepts indexed by $I$.
The impact of this strategy on the number of deterministic RSs is given by the following result:

\begin{proposition}
    \label{prop:concept-supervision}
    Assume that concept supervision is available for all $\vg$ in $\calS \subseteq \mathsf{supp}(\vG)$.
    Under \textbf{A1}, the number of deterministic optima $p_\theta(\vC \mid \vG)$ minimizing the cross-entropy over the concepts is:
    \[
        \label{eq:mc-concept-supervision}
        \textstyle
        \sum_{\alpha \in \calA} \Ind{ \bigwedge_{\vg \in \calS} \bigwedge_{i \in I}  \alpha_i(\vg) = g_i }
    \]
\end{proposition}

This strategy is very powerful:  if $I = [k]$, $\calS \equiv \mathsf{supp}(\vG)$, and the support is complete, there exists only \textit{one} map $\alpha$ that is consistent with the condition in \cref{eq:mc-concept-supervision} and it is the identity.  Naturally, this comes at the cost of obtaining dense annotations for all examples, which is often impractical.

\subsection{Objective-based Mitigation}

A natural alternative is to augment the log-likelihood with an \textit{unsupervised} penalty designed to improve concept quality.
We focus on reconstruction penalties like those used in auto-encoders \citep{hinton1993autoencoders, rezende2014stochastic, kingma2014auto, ghosh2020variational, misino2022vael}, which encourage the model to capture all information necessary to reconstruct the input $\vx$.
To see why these might be useful, consider \cref{ex:mnist-addition-with-shortcuts}.  Here, the model learns a RS mapping both \MZero and \MTwo to the digit $1$:  this RS hinders reconstruction of the input images, and therefore could be avoided by introducing a reconstruction penalty.

In order to implement this, we introduce additional latent variables $\vZ$ that capture the style $\vS$ of the input $\vX$ and modify the concept extractor to output both $\vC$ and $\vZ$, that is:
$
    p_\theta(\vc, \vz \mid \vx) = p_\theta (\vc \mid \vx) \cdot p_\theta(\vz \mid \vx)
$.
The auto-encoder reconstruction penalty is then given by:
\[
    \textstyle
    \calR (\vx)
        = - \bbE_{(\vc, \vz) \sim p_\theta(\vc, \vz \mid \vx) } \big[
                \log p_\psi(\vx \mid \vc, \vz)
            \big]
    \label{eq:rec-loss}
\]
where, $p_\psi(\vx \mid \vc, \vz)$ is the decoder network.  We need to introduce an additional assumption \textbf{A3}:  the encoder and the decoder separate content from style, that is, $p_\theta(\vC, \vZ \mid \vG, \vS) := \bbE_{\vx \sim p^*(\vx \mid \vG, \vS)} p_\theta(\vC, \vZ \mid \vx) $ factorizes as $p_\theta(\vC \mid \vG) p_\theta(\vZ \mid \vS)$ and $p_\psi(\vG, \vS \mid \vC, \vZ) := \bbE_{\vx \sim p_\psi(\vx \mid \vC, \vZ) } p^*(\vG, \vZ \mid \vx)$ as $p_\psi(\vG \mid \vC) p_\psi(\vS \mid \vZ)$.
In this case, we have the following result:

\begin{proposition}
    \label{prop:recon-loss}
    Under \textbf{A1} and \textbf{A3}, the number of deterministic distributions $p_\theta(\vC \mid \vG)$ that minimize the reconstruction penality in \cref{eq:rec-loss} is:
    \[
        \textstyle
        \sum_{\alpha \in \calA} \Ind{
            \bigwedge_{\vg, \vg' \in \mathsf{supp}(\vG) : \vg \ne \vg'}
                \alpha(\vg) \neq \alpha(\vg')
        }
    \]
\end{proposition}

In words, this shows that indeed optimizing for reconstruction facilitates disambiguating between different concepts, \ie different ground-truth concepts cannot be mapped to the same concept. However, minimizing the reconstruction can be non-trivial in practice,
especially for complex inputs.

\subsection{Architecture-based Mitigation}

One last factor is the \textit{architecture of the concept extractor} $p_\theta(\vC \mid \vX)$, as it implicitly controls the number of candidate deterministic maps $\calA$ and therefore the sum in \cref{thm:mc-det-opts}.
If the architecture is unrestricted, $p_\theta(\vC \mid \vX)$ can in principle map any ground-truth concept $\vg$ that generated $\vx$ to any concept $\vc$, thus the cardinality of $\calA$ increases exponentially with $k$.

A powerful strategy for reducing the size of $\calA$ is \textit{disentanglement}.  A model is disentangled if and only if $p_\theta(\vC \mid \vG)$ factorizes as $\prod_{j \in [k]} p_\theta(C_j \mid G_j)$ \citep{locatello2019challenging, suter2019robustly}.
In this case, the maps $\alpha$ also factorize into per-concept maps $\alpha_j: [m_j] \to [m_j]$, dramatically reducing the cardinality of $\calA$, as shown by our first experiment.
In applications where the $k$ concepts are naturally independent from one another, \eg digits in \MNISTAdd, one can implement disentanglement by predicting each concept using the same neural network, although more general techniques exist \citep{locatello2020weakly, shu2019weakly}.

\subsection{Other Heuristics based on Entropy Regularization}

Besides these mitigation strategies, we investigate empirically the effect of the Shannon entropy loss, defined as
$
    1 - \frac{1}{k} \sum_{i=1}^k \Ent_{m_i}[ p_\theta (c_i) ]
$,
which was shown to increase concept quality in DPL \citep{manhaeve2021neural}.
Here, $p_\theta(\vC)$ is the marginal distribution over the concepts, and $\Ent_{m_i}$ is the normalized Shannon entropy over $m_i$ possible values for the distribution. Notice that this term goes to zero only when each distribution $p_\theta(C_i)$ is uniform, which may conflict with the real objective of the NeSy prediction (especially when only few concepts are observed).
\changed{Other similar heuristics that are suited for reducing over-confidence in label predictions are based on label smoothing \citep{muller2019does}, energy-based models \citep{li2021energy}, annealing \citep{li2020closed}, and many others \citep{wei2022mitigating, mukhoti2020calibrating, carratino2022mixup}. In principle, when applied at the concept level, they help reduce the over-confidence in concepts that is typical in deterministic RSs. While they could also be beneficial to mildly reduce the number of RSs, we take the Shannon entropy regularization as a representative of this family for our experiments.}

\section{Case Studies}
\label{sec:experiments}

In this section, we evaluate the impact of RSs in synthetic and real-world NeSy prediction tasks and how the mitigation strategies discussed in \cref{sec:mitigation-strategies} fare in practice.
More details about the models and data are reported in \cref{sec:app-implementations}.
The \textbf{code} is available at \href{https://github.com/ema-marconato/reasoning-shortcuts}{github.com/reasoning-shortcuts}.

\begin{table}
    \centering
    \begin{minipage}{.38\textwidth}
        \scriptsize
%
%
%
%
%
%
\setlength{\tabcolsep}{3pt}
\begin{tabular}{lcccccc}
    \toprule
    & \multicolumn{3}{c}{\XOR}                   & \multicolumn{3}{c}{\MNISTAdd}
    \\
    \cmidrule(lr){2-4} \cmidrule(lr){5-7} 
                                & \multicolumn{1}{c}{DPL}   &  \multicolumn{1}{c}{SL}                         & \multicolumn{1}{c}{LTN}  &  \multicolumn{1}{c}{DPL}                           & \multicolumn{1}{c}{SL}                           & \multicolumn{1}{c}{LTN}
    \\ 
    \rowcolor[HTML]{EFEFEF} --    & $100\%$                   &  $100\%$ &  $100\%$ &  $96.7\%$ &  $82.9\%$ &  $100\%$
    \\
    \DIS    &  $0\%$ &   $0\%$ &   $0\%$ &   $0\%$ &  $0\%$ & $0\%$  
    \\
\bottomrule
\end{tabular}
    \label{tab:dis-XOR}
    \end{minipage}\hfill\begin{minipage}{.55\textwidth}
        \caption{\textbf{Q1: Disentanglement (\DIS) can be very powerful}  to lower the frequency of RSs on \XOR and \MNISTAdd data sets (the lower the better).
        Results are averaged over $30$ \textit{optimal} runs.}
    \end{minipage}
\end{table}

\textbf{Q1: More data does not prevent RSs but disentanglement helps.} We start by evaluating the robustness of DPL, SL, and LTN to RSs in two settings where the data set is \textit{exhaustive}.
In \underline{\XOR} (cf. \cref{ex:xor}), the goal is to predict the parity of three binary concepts $\vg \in \{0, 1\}^3$ given prior knowledge $\BK = (y = g_1 \oplus g_2 \oplus g_3)$.  The predictor receives the ground-truth concepts $\vg$ as input and has to learn a distribution $p_\theta(\vC \mid \vG)$.
In \underline{\MNISTAdd} (cf. \cref{ex:mnist-addition}) the goal is to correctly predict the sum of two MNIST digits, \eg $\vx = (\MOne, \MTwo)$ and $y=3$. 
In both tasks, the training set contains examples of \textit{all} possible combinations of concepts, \ie $\mathsf{supp}(\vG) = \calG$.

Since the tasks are relatively simple, we can afford to study models that achieve near-optimal likelihood, for which \cref{def:reasoning-shortcut} approximately applies.
To this end, for each NeSy architecture, we train several models using different seeds and stop as soon as we obtain $30$ models with likelihood $\ge 0.95$.  Then, we measure the percentage of models that have acquired a RS.  We do the same also for models modified to ensure they are disentangled (\DIS in the Table), see \cref{sec:app-implementations} for details.
The results, reported in \textcolor{blue}{Table~2}, clearly show that RSs affect \textit{all} methods if disentanglement is not in place.  This confirms that, 
without implicit architectural biases,  
\textit{optimizing for label accuracy alone is not sufficient to rule out RSs, even when the data is exhaustive}.
However, when forcing disentanglement, the percentage of models affected by RSs drops to \textit{zero} for all data sets and methods, indicating that mitigation strategies that go beyond the standard learning setup -- and specifically, disentanglement -- can be extremely effective.

\begin{table}
    \begin{minipage}{.25\textwidth}
        \caption{\textbf{Q2: Impact of mitigation strategies.} We report the $F_1$-score on the labels ($\vY$) and concepts ($\vC$). All tested methods incorporate \DIS and are averaged over $10$ runs. \textbf{Top}: NeSy methods combined with \MSR, \MSC, \MSH on \MNISTShortcut. \textbf{Bottom}: evaluation on single tasks \textit{vs} \MTL on \MNISTOp.}
    \end{minipage}\hfill\begin{minipage}{.72\textwidth}
        \centering
        \scriptsize
        \setlength{\tabcolsep}{3pt}
\scalebox{.9}{
\begin{tabular}{lcccccc}
    \toprule
\MNISTShortcut & \multicolumn{2}{c}{DPL}           & \multicolumn{2}{c}{SL}            & \multicolumn{2}{c}{LTN} \\ \cmidrule(lr){2-3} \cmidrule(lr){4-5} \cmidrule(lr){6-7} 
               & $F_1$ ($\vY$)           & $F_1$ ($\vC$)         & $F_1$ ($\vY$)          & $F_1$($\vC$)           & $F_1$ ($\vY$)           & $F_1$($\vC$)                                   \\
\rowcolor[HTML]{EFEFEF}--             & $85.1 \pm 4.6$   & $0.1 \pm 0.1$  & $99.3 \pm 0.2$  & $0.2 \pm 0.1$   &  $ 98.1 \pm 0.2$ & $0.3 \pm 0.1$   \\
\MSR           & $79.8 \pm 1.0$   & $0.1 \pm 0.0$  & $99.5 \pm 0.2$  & $0.1 \pm 0.0$   & $76.3 \pm 1.1$   & $0.0 \pm 0.0$   \\
\rowcolor[HTML]{EFEFEF}\MSH           & $98.1 \pm 0.1$   & $0.1 \pm 0.1$  &  $99.4 \pm 0.1$ & $0.1 \pm 0.0$   & $81.9 \pm 0.5$   & $53.9 \pm 0.7$  \\
\MSC           & ${84.9 \pm 0.1}$ & $0.1 \pm 0.1$  & $99.3 \pm 0.4$  & $21.5 \pm 6.2$  & ${98.1 \pm 0.2}$ & $0.2 \pm 0.1$   \\
\rowcolor[HTML]{EFEFEF}\MSR + \MSH    & ${75.4 \pm 0.4}$ & $0.2 \pm 0.1$  & $99.6 \pm 0.1$  & $0.1 \pm 0.0$   &  $97.9 \pm 2.3$  & $38.1 \pm 16.7$  \\
\MSR + \MSC    & ${84.0 \pm 2.2}$ & $1.9 \pm 4.4$  & $99.3 \pm 0.2$  & $61.5 \pm 7.8$  & ${ 98.1 \pm 0.2 }$ & $0.2 \pm 0.1$ \\
\rowcolor[HTML]{EFEFEF}\MSH + \MSC    & $91.9 \pm 3.5$   & $88.0 \pm 6.3$ &  $99.4 \pm 0.2$ & $41.5 \pm 8.2$  & $98.2 \pm 0.2$   & $98.6 \pm 0.1$ \\
\MSR + \MSH + \MSC & $95.4 \pm 0.4$ & $96.2 \pm 0.2$ & $99.5 \pm 0.2$ & $47.2 \pm 9.8$ & $98.1 \pm 0.3$   & $98.5 \pm 0.2$  \\
\bottomrule \\
\toprule
\MNISTOp       & \multicolumn{2}{c}{DPL}           & \multicolumn{2}{c}{SL}            & \multicolumn{2}{c}{LTN}  \\ \cmidrule(lr){2-3} \cmidrule(lr){4-5} \cmidrule(l){6-7} 
               & $F_1$ ($\vY$)          & $F_1$ ($\vC$)          & $F_1$ ($\vY$)          & $F_1$ ($\vC$)          & $F_1$ ($\vY$)          & $F_1$ ($\vC$) \\
\rowcolor[HTML]{EFEFEF}\textsc{Add}   & $68.1 \pm 6.7$  & $0.0 \pm 0.0$   &  $99.5 \pm 0.2$ & $0.0 \pm 0.1$   & $67.4 \pm 0.1$  & $0.0 \pm 0.0$  \\
\textsc{Mult}  & $100.0 \pm 0.0$ & $37.6 \pm 0.2$  & $100.0 \pm 0.0$ & $76.1 \pm 11.7$ & $98.1 \pm 0.5$  & $78.1 \pm 0.4$ \\
\rowcolor[HTML]{EFEFEF}\textsc{MultiOp}& $100.0 \pm 0.0$ & $99.8 \pm 0.1$ &  $100.0 \pm 0.0$ & $99.8 \pm 0.1$ &  \bf $98.3 \pm 0.2$ &  \bf $98.3 \pm 0.2$ \\ \bottomrule
\end{tabular}
}

        \label{tab:results-MNIST}
    \end{minipage}
\end{table}

\textbf{Q2: Disentanglement is not enough under selection bias.} Next, we look at two (non-exhaustive) data sets where RS occur due to \textit{selection bias}.
Label and concept quality is measured using the $F_1$-score on the macro average measured on the test set.
From here onward, all models are disentangled by construction, and despite this are affected by RSs, as shown below.

We start by evaluating \underline{\MNISTShortcut}, a variant of \MNISTAdd (inspired by \citep{marconato2023neuro}) where only $16$ possible pairs of digits out of $100$ are given for training: $8$ comprise only even digits and $8$ only odd digits.
As shown in \citep{marconato2023neuro}, this setup allows to exchange the semantics of the even and odd digits while ensuring all sums are correct (because, \eg $\MFour + \MEight = 12 = \MNine + \MThree$).
As commonly done in NeSy, hyperparameters were chosen to optimize for label prediction performance on a validation set, cf. \cref{sec:app-implementations}. 
The impact of reconstruction (\MSR), concept supervision (\MSC), and Shannon entropy loss (\MSH) on all architectures are reported in \textcolor{blue}{Table~3}. 
Roughly speaking, a concept $F_1$ below $95\%$ typically indicates a RSs, as shown by the concept confusion matrices in \cref{sec:confusion-matrices}.
The main take away is that \textit{no strategy alone can effectively mitigate RSs for any of the methods}. 
%
Additionally, for DPL and LTN, these also tend to interfere with label supervision, yielding degraded prediction performance.  The SL is not affected by this, likely because it is the only method using a separate neural layer to predict the labels.
Combining multiple strategies does improve concept quality on average, depending on the method.  In particular, $\MSH+\MSC$ and $\MSR+\MSH+\MSC$ help LTN identify good concepts, and similarly for DPL, although concept quality is slightly less stable.
For the SL only concept supervision is relatively effective, but the other strategies are not, probably due to the extra flexibility granted by the top neural layer compared to DPL and LTN.

Next, we evaluate the impact of multi-task learning on an arithmetic task, denoted \underline{\MNISTOp}.  Here, the model observes inputs $\vx \in \{ (\MZero, \MOne), (\MZero, \MTwo), (\MOne, \MThree) \} $ and has to predict both their sum \textit{and} their product, either separately (no \MTL) or jointly (\MTL).
The results in \textcolor{blue}{Table~3} show that, as in \cref{ex:mnist-addition-with-shortcuts}, all methods are dramatically affected by RSs when \MTL is not in place.
DPL and LTN also yield sub-par $F_1$-scores on the labels for the addition task. For SL and LTN, we also observe that, despite high concept $F_1$ for the multiplication task, the digit $\MTwo$ is \textit{never} predicted correctly.  This is clearly visible in the confusion matrices in \cref{sec:confusion-matrices}. 
However, solving addition and multiplication jointly via \MTL ensures all methods acquire very high quality concepts.

\textbf{Q3: Reasoning Shortcuts are pervasive in real-world tasks.} Finally, we look at RSs occurring in \underline{\BOIA} \citep{xu2020boia}, an autonomous vehicle prediction task.
The goal is to predict multiple possible actions $\vY = (\texttt{move\_forward}, \texttt{stop}, \texttt{turn\_left}, \texttt{turn\_right})$ from frames $\vx$ of real driving scenes.
Each scene is described by $21$ binary concepts $\vC$ and the knowledge $\BK$ ensures the predictions are consistent with safety constraints (\eg $\texttt{stop} \Rightarrow \lnot \texttt{move\_forward}$) and guarantees concepts are predicted consistently with one another (\eg $ \texttt{road\_clear} \Leftrightarrow \lnot \texttt{obstacle}$).
%
%
Since this is a high-stakes task, we solve it with DPL, as it is the only approach out of the ones we consider that \textit{guarantees} hard compliance with the knowledge \citep{manhaeve2018deepproblog}.
See \cref{sec:app-implementations} for the full experimental setup.

\cref{tab:result-bdd} lists the results for DPL paired with different mitigation strategies.
In order to get a sense of the model's ability to acquire good concepts, 
\changed{we also include two baselines: {\tt CBM-AUC} introduced by \citet{sawada2022concept} for this task, which learns jointly the concepts together with unsupervised ones, and processes both of them with a linear layer to obtain the labels; 
}
\MSC-only, in which the concept extractor is trained with full concept supervision and frozen, and DPL is only used to perform inference.  This avoids any interference between label and concept supervision.
The $F_1$-scores are computed over the test set and averaged over all $4$ labels and all $21$ concepts.
On the right, we also report the concept confusion matrices (CMs) for the training set:  each row corresponds to a ground-truth concept \textit{vector} $\vg$, and each column to a predicted concept \textit{vector} $\vc$.

Overall, the results show that, unless supplied with concept supervision, DPL optimizes for label accuracy ($F_1$(Y)) by leveraging low quality concepts ($F_1$(C)).  This occurs even when it is paired with a Shannon entropy penalty \MSH.
This is especially evident in the CMs, which show that, for certain labels, all ground-truth concepts $\vg$ are mapped a single $\vc$, not unlike what happens in our first experiment for entangled models.
%
Conversely, the concept quality of DPL substantially improves when concept supervision is available, at the cost of a slight degradation in prediction accuracy.  The CMs back up this observation as the learned concepts tend to align much closer to the diagonal.
Only for the \texttt{turn\_left} label concept supervision fails to prevent collapse.  This occurs because annotations for the corresponding concepts are poor:  for a large set of examples, the concepts necessary to predict $\texttt{turn\_left} = 1$ are annotated as negatives, complicating learning.  In practice, this means that all variants of DPL predict those concepts as negative, hindering concept quality.

\begin{table}[!t]
    \caption{\textbf{Q3}.  \textbf{Left}: Means and std. deviations over $10$ runs for DPL paired with different mitigations. \textbf{Right}: Confusion matrices on the training set for DPL alone (\textit{top}) and paired with mitigation strategies (\textit{bottom}) for $\{\texttt{move\_forward,stop,turn\_left,turn\_right}\}$ concept vectors.}
    \centering
\begin{minipage}{0.44\textwidth}
    \centering
    \scriptsize
    \begin{tabular}{lll}
\toprule
        &  \multicolumn{2}{c}{\BOIA}                                        \\ \cline{2-3} 
        &  \multicolumn{1}{c}{F1-mean (Y)} & \multicolumn{1}{c}{F1-mean(C)} \\
{\tt CBM-AUC}  & $70.8 \pm 0.1$ & $62.1 \pm 0.1$ \\
\rowcolor[HTML]{EFEFEF}\MSC-only &  $64.8 \pm 0.2$               & $60.3 \pm 0.1$              \\
DPL &  $71.4 \pm 0.1$                                       & $39.4 \pm 6.2$              \\
\rowcolor[HTML]{EFEFEF}DPL+\MSH & $\bf 72.1 \pm 0.1$               & $48.1 \pm 0.3$              \\
DPL+\MSC &  $68.2 \pm 0.2$               & $60.5 \pm 0.1$              \\
\rowcolor[HTML]{EFEFEF}DPL+\MSC+\MSH & $68.3 \pm 0.3$               & $\bf 61.7 \pm 0.1$              \\
\bottomrule
\end{tabular}

\end{minipage}
\begin{minipage}{0.54\textwidth}
    \centering
    \scriptsize
    \includegraphics[width=0.95\textwidth]{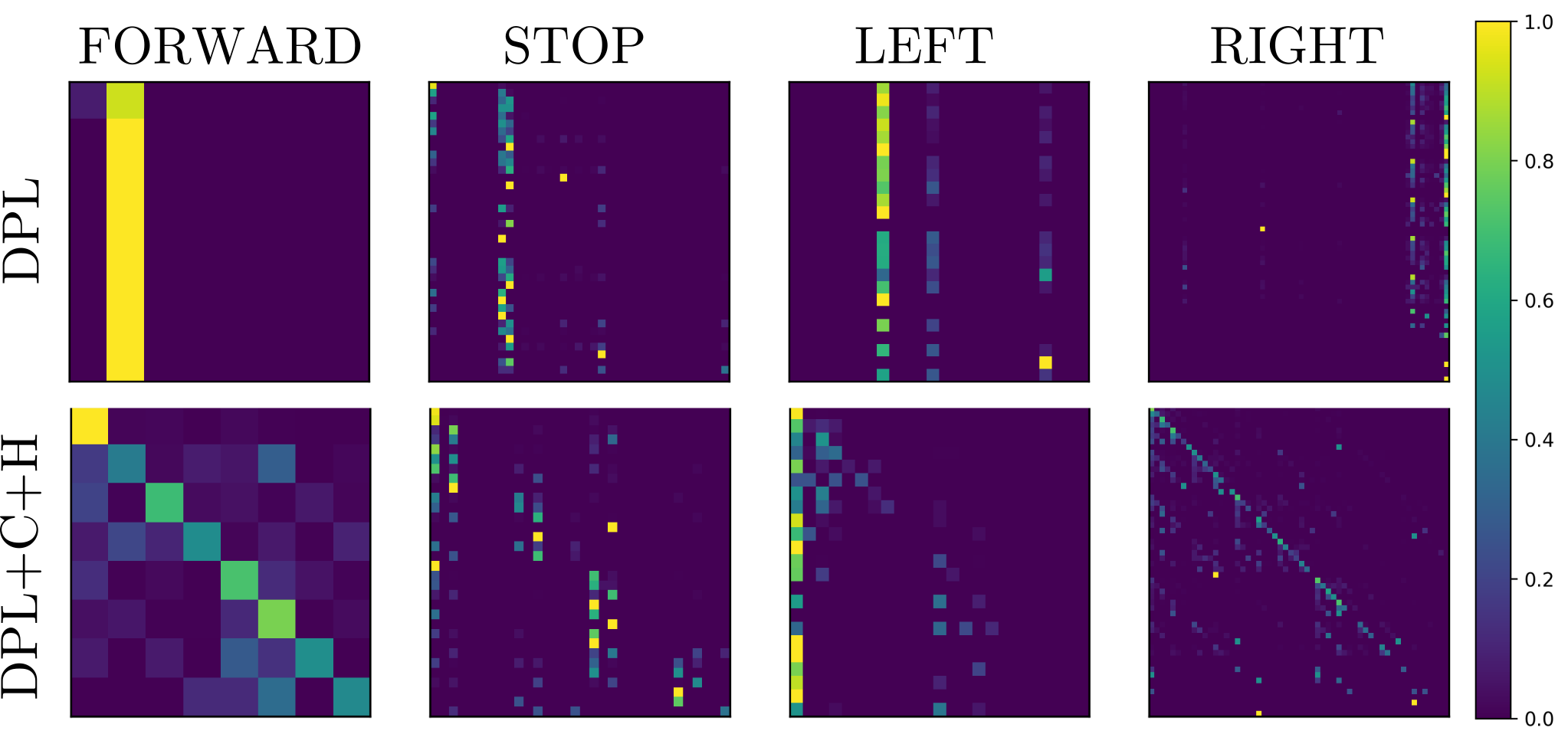}
\end{minipage}
\label{tab:result-bdd}
\end{table}

\section{Related Work}
\label{sec:related-work}

\paragraph{Shortcuts in ML.}  State-of-the-art ML predictors often achieve high performance by exploiting spurious correlations -- or ``shortcuts'' -- in the training data \citep{geirhos2020shortcut}.  Well known examples include watermarks \citep{lapuschkin2019unmasking}, background pixels \citep{xiao2020noise, schramowski2020making}, and textual meta-data in X-ray scans \citep{degrave2021ai}.
Like RSs, regular shortcuts can compromise the classifier's reliability and out-of-distribution generalization and are hard to identify based on accuracy alone.
Proposed solutions include dense annotations~\citep{ross2017right}, out-of-domain data~\citep{parascandolo2020learning}, and interaction with annotators~\citep{teso2023leveraging}.
Shortcuts are often the result of confounding resulting from, \eg selective sampling.  RSs may also arise due to confounding, as is the case in \MNISTAdd (cf. \cref{ex:mnist-addition}), however -- as discussed in \cref{sec:mitigation-strategies},  data is not the only factor underlying RSs.  For instance, in \XOR (cf. \cref{ex:xor}) RSs arise despite exhaustive data.

\textbf{Reasoning shortcuts.}  The issue of RSs has so far been mostly neglected in the NeSy literature, and few remedies have been introduced but never theoretically motivated.
%
%
\citet{stammer2021interactive} have investigated shortcuts affecting NeSy architectures, but consider only \textit{input-level} shortcuts that occur even if concepts are high-quality and fix by injecting additional knowledge in the model.  In contrast, we focus on RSs that impact the quality of learned concepts.
%
%
%
%
\citet{marconato2023neuro} introduce the concept of RSs in the context of NeSy but for continual learning, and proposes a combination of concept supervision and concept-level rehearsal to address them, without delving into a theoretical justification.    
%
%
%
%
%
\citet{li2023learning} propose a minimax objective that ensures the concepts learned by the model satisfy $\BK$.  Like the entropy regularizer \citep{manhaeve2021neural} we addressed in \cref{sec:mitigation-strategies}, this strategy ends up spreading probability across all concepts that satisfy the knowledge, including those that have unintended semantics.  As such, it does not directly address RSs.

\textbf{Neuro-symbolic integration.}  While RSs affect a number of NeSy predictors, NeSy encompasses a heterogeneous family of architectures integrating learning and reasoning~\citep{de2021statistical,garcez2022neural}.
We conjecture that RSs do transfer to \textit{all} NeSy approaches that do not specifically address the factors we identified, but an in-depth analysis of RSs in NeSy is beyond the scope of this paper.

\changed{
\textbf{Relation to disentanglement.} Recovering the latent variables $\vG$ has been the center of several works in representation learning \citep{bengio2013representation, scholkopf2021toward, hyvarinen2023nonlinear}. Among these, achieving identifiability of the latent components has been studied in non-linear independent component analysis \citep{khemakhem2020variational, buchholz2022function, gresele2021independent} and recently in Causal Representation Learning \citep{locatello2020weakly,  ahuja2023interventional, liang2023causal, von2023nonparametric}. In this respect, several overlaps exist with the natural mitigation strategies that we investigated: (i) multi-task learning has been shown to increase disentanglement \citep{maziarka2022relationship} and provably leads to identifiability \citep{lachapelle2023synergies, fumero2023leveraging}, and (ii) latent variables supervision also largely increases the amount of disentanglement \citep{locatello2020disentangling, nie2020semi, shu2019weakly}. Our work is the witness that the intersection with \textit{disentanglement} literature is beneficial for learning the intended concepts in NeSy and, vice-versa, that knowledge-guided learning can be a new way of acquiring identifiable representations, by avoiding RSs. 
}

\section{Conclusion}
\label{sec:conclusion}

In this work, we provide the first in-depth analysis of RS affecting NeSy predictors.  Our analysis highlights four key causes of RS and suggests several mitigation strategies, which we analyze both theoretically and empirically.
Our experiments indicate that RSs do naturally appear in both synthetic and real-world NeSy prediction tasks, and that the effectiveness of mitigation strategies is model and task dependent, and that a general recipe for avoiding RSs is currently missing.

\changed{We foresee that reasoning shortcuts extend beyond the current scope of NeSy predictors with known prior knowledge. This includes approaches that learn jointly both the concepts and the knowledge, like ROAP \citep{tang2023perception} and DSL \citep{daniele2022deep}, as well as fully neural models, like CBMs \citep{koh2020concept} and their variants \citep{marconato2022glancenets, zarlenga2022concept}, which are also likely to be affected by RSs when concept supervision is limited. We plan to investigate further this direction in the near future.}

Ultimately, this work aims at jumpstarting research on the analysis and mitigation of RSs, with the hope of leading to more trustworthy and explainable NeSy architectures.


\textbf{Broader impact.}  Our work brings the subtle but critical issue of RSs to the spotlight, and highlights benefits and limitations of a variety of mitigation strategies, but it is otherwise fundamental research that has no direct societal impact.


\begin{ack}
\changed{We thank the anonymous reviewers for their valuable feedback in improving the current manuscript.} We acknowledge Emile von Krieken for proofreading our claims and Yoshihide Sawada for his help in deploying the models on \BOIA.
We are also grateful to Pedro Zuidberg dos Martires for insightful discussions with us at the early stages of the work.
We acknowledge the support  of the PNRR project FAIR - Future AI Research (PE00000013),  under the NRRP MUR
program funded by the NextGenerationEU. The research of ST and AP was partially supported by TAILOR, a project funded by EU Horizon 2020 research and innovation programme under GA No 952215. 
\end{ack}

\bibliographystyle{unsrtnat}
\bibliography{references, explanatory-supervision}

\begin{thebibliography}{87}
\providecommand{\natexlab}[1]{#1}
\providecommand{\url}[1]{\texttt{#1}}
\expandafter\ifx\csname urlstyle\endcsname\relax
  \providecommand{\doi}[1]{doi: #1}\else
  \providecommand{\doi}{doi: \begingroup \urlstyle{rm}\Url}\fi

\bibitem[De~Raedt et~al.(2021)De~Raedt, Duman{\v{c}}i{\'c}, Manhaeve, and
  Marra]{de2021statistical}
Luc De~Raedt, Sebastijan Duman{\v{c}}i{\'c}, Robin Manhaeve, and Giuseppe
  Marra.
\newblock From statistical relational to neural-symbolic artificial
  intelligence.
\newblock In \emph{Proceedings of the Twenty-Ninth International Conference on
  International Joint Conferences on Artificial Intelligence}, pages
  4943--4950, 2021.

\bibitem[Garcez et~al.(2022)Garcez, Bader, Bowman, Lamb, de~Penning, Illuminoo,
  Poon, and Gerson~Zaverucha]{garcez2022neural}
Artur~d’Avila Garcez, Sebastian Bader, Howard Bowman, Luis~C Lamb, Leo
  de~Penning, BV~Illuminoo, Hoifung Poon, and COPPE Gerson~Zaverucha.
\newblock Neural-symbolic learning and reasoning: A survey and interpretation.
\newblock \emph{Neuro-Symbolic Artificial Intelligence: The State of the Art},
  342:\penalty0 1, 2022.

\bibitem[Giunchiglia et~al.(2022)Giunchiglia, Stoian, and
  Lukasiewicz]{giunchiglia2022deep}
Eleonora Giunchiglia, Mihaela~Catalina Stoian, and Thomas Lukasiewicz.
\newblock Deep learning with logical constraints.
\newblock \emph{arXiv preprint arXiv:2205.00523}, 2022.

\bibitem[Dash et~al.(2022)Dash, Chitlangia, Ahuja, and
  Srinivasan]{dash2022review}
Tirtharaj Dash, Sharad Chitlangia, Aditya Ahuja, and Ashwin Srinivasan.
\newblock A review of some techniques for inclusion of domain-knowledge into
  deep neural networks.
\newblock \emph{Scientific Reports}, 12\penalty0 (1):\penalty0 1--15, 2022.

\bibitem[Diligenti et~al.(2017)Diligenti, Gori, and
  Sacca]{diligenti2017semantic}
Michelangelo Diligenti, Marco Gori, and Claudio Sacca.
\newblock Semantic-based regularization for learning and inference.
\newblock \emph{Artificial Intelligence}, 2017.

\bibitem[Donadello et~al.(2017)Donadello, Serafini, and
  Garcez]{donadello2017logic}
Ivan Donadello, Luciano Serafini, and Artur~D'Avila Garcez.
\newblock Logic tensor networks for semantic image interpretation.
\newblock In \emph{IJCAI}, 2017.

\bibitem[Manhaeve et~al.(2018)Manhaeve, Dumancic, Kimmig, Demeester, and
  De~Raedt]{manhaeve2018deepproblog}
Robin Manhaeve, Sebastijan Dumancic, Angelika Kimmig, Thomas Demeester, and Luc
  De~Raedt.
\newblock {DeepProbLog: Neural Probabilistic Logic Programming}.
\newblock \emph{NeurIPS}, 2018.

\bibitem[Xu et~al.(2018)Xu, Zhang, Friedman, Liang, and Broeck]{xu2018semantic}
Jingyi Xu, Zilu Zhang, Tal Friedman, Yitao Liang, and Guy Broeck.
\newblock A semantic loss function for deep learning with symbolic knowledge.
\newblock In \emph{ICML}, 2018.

\bibitem[Giunchiglia and Lukasiewicz(2020)]{giunchiglia2020coherent}
Eleonora Giunchiglia and Thomas Lukasiewicz.
\newblock Coherent hierarchical multi-label classification networks.
\newblock \emph{NeurIPS}, 2020.

\bibitem[Ahmed et~al.(2022{\natexlab{a}})Ahmed, Teso, Chang, Van~den Broeck,
  and Vergari]{ahmed2022semantic}
Kareem Ahmed, Stefano Teso, Kai-Wei Chang, Guy Van~den Broeck, and Antonio
  Vergari.
\newblock {Semantic Probabilistic Layers for Neuro-Symbolic Learning}.
\newblock In \emph{NeurIPS}, 2022{\natexlab{a}}.

\bibitem[Li et~al.(2023)Li, Liu, Yao, Xu, Chen, Ma, Jian,
  et~al.]{li2023learning}
Zenan Li, Zehua Liu, Yuan Yao, Jingwei Xu, Taolue Chen, Xiaoxing Ma, L~Jian,
  et~al.
\newblock Learning with logical constraints but without shortcut satisfaction.
\newblock In \emph{The Eleventh International Conference on Learning
  Representations}, 2023.

\bibitem[Xie et~al.(2022)Xie, Kersting, and Neider]{xie2022neuro}
Xuan Xie, Kristian Kersting, and Daniel Neider.
\newblock Neuro-symbolic verification of deep neural networks.
\newblock 2022.

\bibitem[Rudin(2019)]{rudin2019stop}
Cynthia Rudin.
\newblock Stop explaining black box machine learning models for high stakes
  decisions and use interpretable models instead.
\newblock \emph{Nature Machine Intelligence}, 1\penalty0 (5):\penalty0
  206--215, 2019.

\bibitem[Chen et~al.(2019)Chen, Li, Tao, Barnett, Rudin, and Su]{chen2019looks}
Chaofan Chen, Oscar Li, Daniel Tao, Alina Barnett, Cynthia Rudin, and
  Jonathan~K Su.
\newblock This looks like that: Deep learning for interpretable image
  recognition.
\newblock \emph{NeurIPS}, 2019.

\bibitem[Chen et~al.(2020)Chen, Bei, and Rudin]{chen2020concept}
Zhi Chen, Yijie Bei, and Cynthia Rudin.
\newblock Concept whitening for interpretable image recognition.
\newblock \emph{Nature Machine Intelligence}, 2020.

\bibitem[DeGrave et~al.(2021)DeGrave, Janizek, and Lee]{degrave2021ai}
Alex~J DeGrave, Joseph~D Janizek, and Su-In Lee.
\newblock Ai for radiographic covid-19 detection selects shortcuts over signal.
\newblock \emph{Nature Machine Intelligence}, pages 1--10, 2021.

\bibitem[Maiettini et~al.(2019)Maiettini, Pasquale, Tikhanoff, Rosasco, and
  Natale]{maiettini2019weakly}
Elisa Maiettini, Giulia Pasquale, Vadim Tikhanoff, Lorenzo Rosasco, and Lorenzo
  Natale.
\newblock A weakly supervised strategy for learning object detection on a
  humanoid robot.
\newblock In \emph{2019 IEEE-RAS 19th International Conference on Humanoid
  Robots (Humanoids)}, pages 194--201. IEEE, 2019.

\bibitem[Badue et~al.(2021)Badue, Guidolini, Carneiro, Azevedo, Cardoso,
  Forechi, Jesus, Berriel, Paixao, Mutz, et~al.]{badue2021self}
Claudine Badue, R{\^a}nik Guidolini, Raphael~Vivacqua Carneiro, Pedro Azevedo,
  Vinicius~B Cardoso, Avelino Forechi, Luan Jesus, Rodrigo Berriel, Thiago~M
  Paixao, Filipe Mutz, et~al.
\newblock Self-driving cars: A survey.
\newblock \emph{Expert Systems with Applications}, 165:\penalty0 113816, 2021.

\bibitem[Fredrikson et~al.(2023)Fredrikson, Lu, Vijayakumar, Jha, Ganesh, and
  Wang]{fredrikson2023learning}
Matt Fredrikson, Kaiji Lu, Saranya Vijayakumar, Somesh Jha, Vijay Ganesh, and
  Zifan Wang.
\newblock Learning modulo theories.
\newblock \emph{arXiv preprint arXiv:2301.11435}, 2023.

\bibitem[Manhaeve et~al.(2021{\natexlab{a}})Manhaeve, Duman{\v{c}}i{\'c},
  Kimmig, Demeester, and De~Raedt]{manhaeve2021neural}
Robin Manhaeve, Sebastijan Duman{\v{c}}i{\'c}, Angelika Kimmig, Thomas
  Demeester, and Luc De~Raedt.
\newblock Neural probabilistic logic programming in deepproblog.
\newblock \emph{Artificial Intelligence}, 298:\penalty0 103504,
  2021{\natexlab{a}}.

\bibitem[Marconato et~al.(2023)Marconato, Bontempo, Ficarra, Calderara,
  Passerini, and Teso]{marconato2023neuro}
Emanuele Marconato, Gianpaolo Bontempo, Elisa Ficarra, Simone Calderara, Andrea
  Passerini, and Stefano Teso.
\newblock Neuro symbolic continual learning: Knowledge, reasoning shortcuts and
  concept rehearsal.
\newblock In \emph{ICML}, 2023.

\bibitem[LeCun(1998)]{lecun1998mnist}
Yann LeCun.
\newblock The mnist database of handwritten digits.
\newblock \emph{http://yann. lecun. com/exdb/mnist/}, 1998.

\bibitem[Fischer et~al.(2019)Fischer, Balunovic, Drachsler-Cohen, Gehr, Zhang,
  and Vechev]{fischer2019dl2}
Marc Fischer, Mislav Balunovic, Dana Drachsler-Cohen, Timon Gehr, Ce~Zhang, and
  Martin Vechev.
\newblock Dl2: Training and querying neural networks with logic.
\newblock In \emph{International Conference on Machine Learning}, pages
  1931--1941. PMLR, 2019.

\bibitem[Ahmed et~al.(2022{\natexlab{b}})Ahmed, Wang, Chang, and Van~den
  Broeck]{ahmed2022neuro}
Kareem Ahmed, Eric Wang, Kai-Wei Chang, and Guy Van~den Broeck.
\newblock Neuro-symbolic entropy regularization.
\newblock In \emph{UAI}, 2022{\natexlab{b}}.

\bibitem[Hoernle et~al.(2022)Hoernle, Karampatsis, Belle, and
  Gal]{hoernle2022multiplexnet}
Nick Hoernle, Rafael~Michael Karampatsis, Vaishak Belle, and Kobi Gal.
\newblock Multiplexnet: Towards fully satisfied logical constraints in neural
  networks.
\newblock In \emph{AAAI}, 2022.

\bibitem[Diligenti et~al.(2012)Diligenti, Gori, Maggini, and
  Rigutini]{diligenti2012bridging}
Michelangelo Diligenti, Marco Gori, Marco Maggini, and Leonardo Rigutini.
\newblock Bridging logic and kernel machines.
\newblock \emph{Machine learning}, 86\penalty0 (1):\penalty0 57--88, 2012.

\bibitem[Pryor et~al.(2022)Pryor, Dickens, Augustine, Albalak, Wang, and
  Getoor]{pryor2022neupsl}
Connor Pryor, Charles Dickens, Eriq Augustine, Alon Albalak, William Wang, and
  Lise Getoor.
\newblock Neupsl: Neural probabilistic soft logic.
\newblock \emph{arXiv preprint arXiv:2205.14268}, 2022.

\bibitem[De~Raedt and Kimmig(2015)]{de2015probabilistic}
Luc De~Raedt and Angelika Kimmig.
\newblock Probabilistic (logic) programming concepts.
\newblock \emph{Machine Learning}, 2015.

\bibitem[Darwiche and Marquis(2002)]{darwiche2002knowledge}
Adnan Darwiche and Pierre Marquis.
\newblock A knowledge compilation map.
\newblock \emph{Journal of Artificial Intelligence Research}, 17:\penalty0
  229--264, 2002.

\bibitem[Vergari et~al.(2021)Vergari, Choi, Liu, Teso, and Van~den
  Broeck]{vergari2021compositional}
Antonio Vergari, YooJung Choi, Anji Liu, Stefano Teso, and Guy Van~den Broeck.
\newblock A compositional atlas of tractable circuit operations for
  probabilistic inference.
\newblock \emph{Advances in Neural Information Processing Systems}, 34, 2021.

\bibitem[Manhaeve et~al.(2021{\natexlab{b}})Manhaeve, Marra, and
  De~Raedt]{manhaeve2021approximate}
Robin Manhaeve, Giuseppe Marra, and Luc De~Raedt.
\newblock Approximate inference for neural probabilistic logic programming.
\newblock In \emph{KR}, 2021{\natexlab{b}}.

\bibitem[Huang et~al.(2021)Huang, Li, Chen, Samel, Naik, Song, and
  Si]{huang2021scallop}
Jiani Huang, Ziyang Li, Binghong Chen, Karan Samel, Mayur Naik, Le~Song, and
  Xujie Si.
\newblock Scallop: From probabilistic deductive databases to scalable
  differentiable reasoning.
\newblock \emph{NeurIPS}, 2021.

\bibitem[Winters et~al.(2022)Winters, Marra, Manhaeve, and
  De~Raedt]{winters2022deepstochlog}
Thomas Winters, Giuseppe Marra, Robin Manhaeve, and Luc De~Raedt.
\newblock {DeepStochLog: Neural Stochastic Logic Programming}.
\newblock In \emph{AAAI}, 2022.

\bibitem[van Krieken et~al.(2022)van Krieken, Thanapalasingam, Tomczak, van
  Harmelen, and Teije]{van2022anesi}
Emile van Krieken, Thiviyan Thanapalasingam, Jakub~M Tomczak, Frank van
  Harmelen, and Annette~ten Teije.
\newblock A-nesi: A scalable approximate method for probabilistic neurosymbolic
  inference.
\newblock \emph{arXiv preprint arXiv:2212.12393}, 2022.

\bibitem[Sch{\"o}lkopf et~al.(2021)Sch{\"o}lkopf, Locatello, Bauer, Ke,
  Kalchbrenner, Goyal, and Bengio]{scholkopf2021toward}
Bernhard Sch{\"o}lkopf, Francesco Locatello, Stefan Bauer, Nan~Rosemary Ke, Nal
  Kalchbrenner, Anirudh Goyal, and Yoshua Bengio.
\newblock Toward causal representation learning.
\newblock \emph{Proceedings of the IEEE}, 2021.

\bibitem[Suter et~al.(2019)Suter, Miladinovic, Sch{\"o}lkopf, and
  Bauer]{suter2019robustly}
Raphael Suter, Djordje Miladinovic, Bernhard Sch{\"o}lkopf, and Stefan Bauer.
\newblock Robustly disentangled causal mechanisms: Validating deep
  representations for interventional robustness.
\newblock In \emph{International Conference on Machine Learning}, pages
  6056--6065. PMLR, 2019.

\bibitem[Von~K{\"u}gelgen et~al.(2021)Von~K{\"u}gelgen, Sharma, Gresele,
  Brendel, Sch{\"o}lkopf, Besserve, and Locatello]{vonkugelgen2021self}
Julius Von~K{\"u}gelgen, Yash Sharma, Luigi Gresele, Wieland Brendel, Bernhard
  Sch{\"o}lkopf, Michel Besserve, and Francesco Locatello.
\newblock Self-supervised learning with data augmentations provably isolates
  content from style.
\newblock \emph{Advances in neural information processing systems},
  34:\penalty0 16451--16467, 2021.

\bibitem[Quinonero-Candela et~al.(2008)Quinonero-Candela, Sugiyama,
  Schwaighofer, and Lawrence]{quinonero2008dataset}
Joaquin Quinonero-Candela, Masashi Sugiyama, Anton Schwaighofer, and Neil~D
  Lawrence.
\newblock \emph{Dataset shift in machine learning}.
\newblock Mit Press, 2008.

\bibitem[Marconato et~al.(2022)Marconato, Passerini, and
  Teso]{marconato2022glancenets}
Emanuele Marconato, Andrea Passerini, and Stefano Teso.
\newblock Glancenets: Interpretabile, leak-proof concept-based models.
\newblock \emph{NeurIPS}, 2022.

\bibitem[Hinton and Zemel(1993)]{hinton1993autoencoders}
Geoffrey~E Hinton and Richard Zemel.
\newblock Autoencoders, minimum description length and helmholtz free energy.
\newblock \emph{Advances in neural information processing systems}, 6, 1993.

\bibitem[Rezende et~al.(2014)Rezende, Mohamed, and
  Wierstra]{rezende2014stochastic}
Danilo~Jimenez Rezende, Shakir Mohamed, and Daan Wierstra.
\newblock Stochastic backpropagation and approximate inference in deep
  generative models.
\newblock In \emph{International conference on machine learning}, 2014.

\bibitem[Kingma and Welling(2014)]{kingma2014auto}
Diederik~P Kingma and Max Welling.
\newblock Auto-encoding variational bayes.
\newblock In \emph{International conference on machine learning}, 2014.

\bibitem[Ghosh et~al.(2020)Ghosh, Sajjadi, Vergari, Black, and
  Sch{\"o}lkopf]{ghosh2020variational}
Partha Ghosh, Mehdi~SM Sajjadi, Antonio Vergari, Michael~J Black, and Bernhard
  Sch{\"o}lkopf.
\newblock From variational to deterministic autoencoders.
\newblock In \emph{ICLR}, 2020.

\bibitem[Misino et~al.(2022)Misino, Marra, and Sansone]{misino2022vael}
Eleonora Misino, Giuseppe Marra, and Emanuele Sansone.
\newblock {VAEL: Bridging Variational Autoencoders and Probabilistic Logic
  Programming}.
\newblock \emph{NeurIPS}, 2022.

\bibitem[Locatello et~al.(2019)Locatello, Bauer, Lucic, Raetsch, Gelly,
  Sch{\"o}lkopf, and Bachem]{locatello2019challenging}
Francesco Locatello, Stefan Bauer, Mario Lucic, Gunnar Raetsch, Sylvain Gelly,
  Bernhard Sch{\"o}lkopf, and Olivier Bachem.
\newblock Challenging common assumptions in the unsupervised learning of
  disentangled representations.
\newblock In \emph{ICML}, 2019.

\bibitem[Locatello et~al.(2020{\natexlab{a}})Locatello, Poole, R{\"a}tsch,
  Sch{\"o}lkopf, Bachem, and Tschannen]{locatello2020weakly}
Francesco Locatello, Ben Poole, Gunnar R{\"a}tsch, Bernhard Sch{\"o}lkopf,
  Olivier Bachem, and Michael Tschannen.
\newblock Weakly-supervised disentanglement without compromises.
\newblock In \emph{International Conference on Machine Learning}, pages
  6348--6359. PMLR, 2020{\natexlab{a}}.

\bibitem[Shu et~al.(2019)Shu, Chen, Kumar, Ermon, and Poole]{shu2019weakly}
Rui Shu, Yining Chen, Abhishek Kumar, Stefano Ermon, and Ben Poole.
\newblock Weakly supervised disentanglement with guarantees.
\newblock In \emph{ICLR}, 2019.

\bibitem[M{\"u}ller et~al.(2019)M{\"u}ller, Kornblith, and
  Hinton]{muller2019does}
Rafael M{\"u}ller, Simon Kornblith, and Geoffrey~E Hinton.
\newblock When does label smoothing help?
\newblock \emph{Advances in neural information processing systems}, 32, 2019.

\bibitem[Li et~al.(2021)Li, Du, van~de Ven, and Mordatch]{li2021energy}
Shuang Li, Yilun Du, Gido~Martijn van~de Ven, and Igor Mordatch.
\newblock Energy-based models for continual learning.
\newblock In \emph{Energy Based Models Workshop-ICLR 2021}, 2021.

\bibitem[Li et~al.(2020)Li, Huang, Hong, Chen, Wu, and Zhu]{li2020closed}
Qing Li, Siyuan Huang, Yining Hong, Yixin Chen, Ying~Nian Wu, and Song-Chun
  Zhu.
\newblock Closed loop neural-symbolic learning via integrating neural
  perception, grammar parsing, and symbolic reasoning.
\newblock In \emph{International Conference on Machine Learning}, pages
  5884--5894. PMLR, 2020.

\bibitem[Wei et~al.(2022)Wei, Xie, Cheng, Feng, An, and Li]{wei2022mitigating}
Hongxin Wei, Renchunzi Xie, Hao Cheng, Lei Feng, Bo~An, and Yixuan Li.
\newblock Mitigating neural network overconfidence with logit normalization.
\newblock In \emph{International Conference on Machine Learning}, pages
  23631--23644. PMLR, 2022.

\bibitem[Mukhoti et~al.(2020)Mukhoti, Kulharia, Sanyal, Golodetz, Torr, and
  Dokania]{mukhoti2020calibrating}
Jishnu Mukhoti, Viveka Kulharia, Amartya Sanyal, Stuart Golodetz, Philip Torr,
  and Puneet Dokania.
\newblock Calibrating deep neural networks using focal loss.
\newblock \emph{Advances in Neural Information Processing Systems},
  33:\penalty0 15288--15299, 2020.

\bibitem[Carratino et~al.(2022)Carratino, Ciss{\'e}, Jenatton, and
  Vert]{carratino2022mixup}
Luigi Carratino, Moustapha Ciss{\'e}, Rodolphe Jenatton, and Jean-Philippe
  Vert.
\newblock On mixup regularization.
\newblock \emph{The Journal of Machine Learning Research}, 23\penalty0
  (1):\penalty0 14632--14662, 2022.

\bibitem[Xu et~al.(2020)Xu, Yang, Gong, Lin, Wu, Li, and
  Vasconcelos]{xu2020boia}
Yiran Xu, Xiaoyin Yang, Lihang Gong, Hsuan-Chu Lin, Tz-Ying Wu, Yunsheng Li,
  and Nuno Vasconcelos.
\newblock Explainable object-induced action decision for autonomous vehicles.
\newblock In \emph{IEEE/CVF Conference on Computer Vision and Pattern
  Recognition (CVPR)}, June 2020.

\bibitem[Sawada and Nakamura(2022)]{sawada2022concept}
Yoshihide Sawada and Keigo Nakamura.
\newblock Concept bottleneck model with additional unsupervised concepts.
\newblock \emph{IEEE Access}, 10:\penalty0 41758--41765, 2022.

\bibitem[Geirhos et~al.(2020)Geirhos, Jacobsen, Michaelis, Zemel, Brendel,
  Bethge, and Wichmann]{geirhos2020shortcut}
Robert Geirhos, J{\"o}rn-Henrik Jacobsen, Claudio Michaelis, Richard Zemel,
  Wieland Brendel, Matthias Bethge, and Felix~A Wichmann.
\newblock Shortcut learning in deep neural networks.
\newblock \emph{Nature Machine Intelligence}, 2\penalty0 (11):\penalty0
  665--673, 2020.

\bibitem[Lapuschkin et~al.(2019)Lapuschkin, W{\"a}ldchen, Binder, Montavon,
  Samek, and M{\"u}ller]{lapuschkin2019unmasking}
Sebastian Lapuschkin, Stephan W{\"a}ldchen, Alexander Binder, Gr{\'e}goire
  Montavon, Wojciech Samek, and Klaus-Robert M{\"u}ller.
\newblock Unmasking clever hans predictors and assessing what machines really
  learn.
\newblock \emph{Nature communications}, 10\penalty0 (1):\penalty0 1--8, 2019.

\bibitem[Xiao et~al.(2020)Xiao, Engstrom, Ilyas, and Madry]{xiao2020noise}
Kai~Yuanqing Xiao, Logan Engstrom, Andrew Ilyas, and Aleksander Madry.
\newblock Noise or signal: The role of image backgrounds in object recognition.
\newblock In \emph{ICLR}, 2020.

\bibitem[Schramowski et~al.(2020)Schramowski, Stammer, Teso, Brugger, Herbert,
  Shao, Luigs, Mahlein, and Kersting]{schramowski2020making}
Patrick Schramowski, Wolfgang Stammer, Stefano Teso, Anna Brugger, Franziska
  Herbert, Xiaoting Shao, Hans-Georg Luigs, Anne-Katrin Mahlein, and Kristian
  Kersting.
\newblock Making deep neural networks right for the right scientific reasons by
  interacting with their explanations.
\newblock \emph{Nature Machine Intelligence}, 2\penalty0 (8):\penalty0
  476--486, 2020.

\bibitem[Ross et~al.(2017)Ross, Hughes, and Doshi-Velez]{ross2017right}
Andrew~Slavin Ross, Michael~C Hughes, and Finale Doshi-Velez.
\newblock Right for the right reasons: training differentiable models by
  constraining their explanations.
\newblock In \emph{Proceedings of the 26th International Joint Conference on
  Artificial Intelligence}, pages 2662--2670, 2017.

\bibitem[Parascandolo et~al.(2020)Parascandolo, Neitz, ORVIETO, Gresele, and
  Sch{\"o}lkopf]{parascandolo2020learning}
Giambattista Parascandolo, Alexander Neitz, ANTONIO ORVIETO, Luigi Gresele, and
  Bernhard Sch{\"o}lkopf.
\newblock Learning explanations that are hard to vary.
\newblock In \emph{International Conference on Learning Representations}, 2020.

\bibitem[Teso et~al.(2023)Teso, Alkan, Stammer, and Daly]{teso2023leveraging}
Stefano Teso, {\"O}znur Alkan, Wolfang Stammer, and Elizabeth Daly.
\newblock Leveraging explanations in interactive machine learning: An overview.
\newblock \emph{Frontiers in Artificial Intelligence}, 2023.

\bibitem[Stammer et~al.(2021)Stammer, Memmel, Schramowski, and
  Kersting]{stammer2021interactive}
Wolfgang Stammer, Marius Memmel, Patrick Schramowski, and Kristian Kersting.
\newblock Interactive disentanglement: Learning concepts by interacting with
  their prototype representations.
\newblock \emph{arXiv preprint arXiv:2112.02290}, 2021.

\bibitem[Bengio et~al.(2013)Bengio, Courville, and
  Vincent]{bengio2013representation}
Yoshua Bengio, Aaron Courville, and Pascal Vincent.
\newblock Representation learning: A review and new perspectives.
\newblock \emph{IEEE transactions on pattern analysis and machine
  intelligence}, 35\penalty0 (8):\penalty0 1798--1828, 2013.

\bibitem[Hyv{\"a}rinen et~al.(2023)Hyv{\"a}rinen, Khemakhem, and
  Morioka]{hyvarinen2023nonlinear}
Aapo Hyv{\"a}rinen, Ilyes Khemakhem, and Hiroshi Morioka.
\newblock Nonlinear independent component analysis for principled
  disentanglement in unsupervised deep learning.
\newblock \emph{Patterns}, 4\penalty0 (10), 2023.

\bibitem[Khemakhem et~al.(2020)Khemakhem, Kingma, Monti, and
  Hyvarinen]{khemakhem2020variational}
Ilyes Khemakhem, Diederik Kingma, Ricardo Monti, and Aapo Hyvarinen.
\newblock {Variational autoencoders and nonlinear ICA: A unifying framework}.
\newblock In \emph{AISTATS}, 2020.

\bibitem[Buchholz et~al.(2022)Buchholz, Besserve, and
  Sch{\"o}lkopf]{buchholz2022function}
Simon Buchholz, Michel Besserve, and Bernhard Sch{\"o}lkopf.
\newblock Function classes for identifiable nonlinear independent component
  analysis.
\newblock \emph{Advances in Neural Information Processing Systems},
  35:\penalty0 16946--16961, 2022.

\bibitem[Gresele et~al.(2021)Gresele, Von~K{\"u}gelgen, Stimper, Sch{\"o}lkopf,
  and Besserve]{gresele2021independent}
Luigi Gresele, Julius Von~K{\"u}gelgen, Vincent Stimper, Bernhard
  Sch{\"o}lkopf, and Michel Besserve.
\newblock Independent mechanism analysis, a new concept?
\newblock \emph{Advances in neural information processing systems},
  34:\penalty0 28233--28248, 2021.

\bibitem[Ahuja et~al.(2023)Ahuja, Mahajan, Wang, and
  Bengio]{ahuja2023interventional}
Kartik Ahuja, Divyat Mahajan, Yixin Wang, and Yoshua Bengio.
\newblock Interventional causal representation learning.
\newblock In \emph{International conference on machine learning}, pages
  372--407. PMLR, 2023.

\bibitem[Liang et~al.(2023)Liang, Keki{\'c}, von K{\"u}gelgen, Buchholz,
  Besserve, Gresele, and Sch{\"o}lkopf]{liang2023causal}
Wendong Liang, Armin Keki{\'c}, Julius von K{\"u}gelgen, Simon Buchholz, Michel
  Besserve, Luigi Gresele, and Bernhard Sch{\"o}lkopf.
\newblock Causal component analysis.
\newblock \emph{arXiv preprint arXiv:2305.17225}, 2023.

\bibitem[von K{\"u}gelgen et~al.(2023)von K{\"u}gelgen, Besserve, Liang,
  Gresele, Keki{\'c}, Bareinboim, Blei, and
  Sch{\"o}lkopf]{von2023nonparametric}
Julius von K{\"u}gelgen, Michel Besserve, Wendong Liang, Luigi Gresele, Armin
  Keki{\'c}, Elias Bareinboim, David~M Blei, and Bernhard Sch{\"o}lkopf.
\newblock Nonparametric identifiability of causal representations from unknown
  interventions.
\newblock \emph{arXiv preprint arXiv:2306.00542}, 2023.

\bibitem[Maziarka et~al.(2022)Maziarka, Nowak, Wo{\l}czyk, and
  Bedychaj]{maziarka2022relationship}
{\L}ukasz Maziarka, Aleksandra Nowak, Maciej Wo{\l}czyk, and Andrzej Bedychaj.
\newblock On the relationship between disentanglement and multi-task learning.
\newblock In \emph{Joint European Conference on Machine Learning and Knowledge
  Discovery in Databases}, pages 625--641. Springer, 2022.

\bibitem[Lachapelle et~al.(2023)Lachapelle, Deleu, Mahajan, Mitliagkas, Bengio,
  Lacoste-Julien, and Bertrand]{lachapelle2023synergies}
S{\'e}bastien Lachapelle, Tristan Deleu, Divyat Mahajan, Ioannis Mitliagkas,
  Yoshua Bengio, Simon Lacoste-Julien, and Quentin Bertrand.
\newblock Synergies between disentanglement and sparsity: generalization and
  identifiability in multi-task learning.
\newblock In \emph{International Conference on Machine Learning}, pages
  18171--18206. PMLR, 2023.

\bibitem[Fumero et~al.(2023)Fumero, Wenzel, Zancato, Achille, Rodol{\`a},
  Soatto, Sch{\"o}lkopf, and Locatello]{fumero2023leveraging}
Marco Fumero, Florian Wenzel, Luca Zancato, Alessandro Achille, Emanuele
  Rodol{\`a}, Stefano Soatto, Bernhard Sch{\"o}lkopf, and Francesco Locatello.
\newblock Leveraging sparse and shared feature activations for disentangled
  representation learning.
\newblock \emph{arXiv preprint arXiv:2304.07939}, 2023.

\bibitem[Locatello et~al.(2020{\natexlab{b}})Locatello, Tschannen, Bauer,
  R{\"a}tsch, Sch{\"o}lkopf, and Bachem]{locatello2020disentangling}
Francesco Locatello, Michael Tschannen, Stefan Bauer, Gunnar R{\"a}tsch,
  Bernhard Sch{\"o}lkopf, and Olivier Bachem.
\newblock Disentangling factors of variations using few labels.
\newblock In \emph{International Conference on Learning Representations},
  2020{\natexlab{b}}.

\bibitem[Nie et~al.(2020)Nie, Karras, Garg, Debnath, Patney, Patel, and
  Anandkumar]{nie2020semi}
Weili Nie, Tero Karras, Animesh Garg, Shoubhik Debnath, Anjul Patney, Ankit~B
  Patel, and Anima Anandkumar.
\newblock Semi-supervised stylegan for disentanglement learning.
\newblock In \emph{Proceedings of the 37th International Conference on Machine
  Learning}, pages 7360--7369, 2020.

\bibitem[Tang and Ellis(2023)]{tang2023perception}
Hao Tang and Kevin Ellis.
\newblock From perception to programs: regularize, overparameterize, and
  amortize.
\newblock In \emph{International Conference on Machine Learning}, pages
  33616--33631. PMLR, 2023.

\bibitem[Daniele et~al.(2022)Daniele, Campari, Malhotra, and
  Serafini]{daniele2022deep}
Alessandro Daniele, Tommaso Campari, Sagar Malhotra, and Luciano Serafini.
\newblock Deep symbolic learning: Discovering symbols and rules from
  perceptions.
\newblock \emph{arXiv preprint arXiv:2208.11561}, 2022.

\bibitem[Koh et~al.(2020)Koh, Nguyen, Tang, Mussmann, Pierson, Kim, and
  Liang]{koh2020concept}
Pang~Wei Koh, Thao Nguyen, Yew~Siang Tang, Stephen Mussmann, Emma Pierson, Been
  Kim, and Percy Liang.
\newblock Concept bottleneck models.
\newblock In \emph{International Conference on Machine Learning}, pages
  5338--5348. PMLR, 2020.

\bibitem[Zarlenga et~al.(2022)Zarlenga, Barbiero, Ciravegna, Marra, Giannini,
  Diligenti, Shams, Precioso, Melacci, Weller, et~al.]{zarlenga2022concept}
Mateo~Espinosa Zarlenga, Pietro Barbiero, Gabriele Ciravegna, Giuseppe Marra,
  Francesco Giannini, Michelangelo Diligenti, Zohreh Shams, Frederic Precioso,
  Stefano Melacci, Adrian Weller, et~al.
\newblock Concept embedding models.
\newblock \emph{arXiv preprint arXiv:2209.09056}, 2022.

\bibitem[Giannini et~al.(2018)Giannini, Diligenti, Gori, and
  Maggini]{giannini2018convex}
Francesco Giannini, Michelangelo Diligenti, Marco Gori, and Marco Maggini.
\newblock On a convex logic fragment for learning and reasoning.
\newblock \emph{IEEE Transactions on Fuzzy Systems}, 2018.

\bibitem[Cover(1999)]{cover1999elements}
Thomas~M Cover.
\newblock \emph{Elements of information theory}.
\newblock John Wiley \& Sons, 1999.

\bibitem[Ahmed et~al.(2023)Ahmed, Chang, and Van~den Broeck]{AhmedKLR23}
Kareem Ahmed, Kai-Wei Chang, and Guy Van~den Broeck.
\newblock A pseudo-semantic loss for deep generative models with logical
  constraints.
\newblock In \emph{Knowledge and Logical Reasoning in the Era of Data-driven
  Learning Workshop}, July 2023.

\bibitem[Paszke et~al.(2019)Paszke, Gross, Massa, Lerer, Bradbury, Chanan,
  Killeen, Lin, Gimelshein, Antiga, et~al.]{paszke2019pytorch}
Adam Paszke, Sam Gross, Francisco Massa, Adam Lerer, James Bradbury, Gregory
  Chanan, Trevor Killeen, Zeming Lin, Natalia Gimelshein, Luca Antiga, et~al.
\newblock Pytorch: An imperative style, high-performance deep learning library.
\newblock \emph{Advances in neural information processing systems}, 32, 2019.

\bibitem[Carraro(2022)]{LTNtorch}
Tommaso Carraro.
\newblock {LTNtorch: PyTorch implementation of Logic Tensor Networks}, mar
  2022.
\newblock URL \url{https://doi.org/10.5281/zenodo.6394282}.

\bibitem[Kingma and Ba(2015)]{KingmaB14@adam}
Diederik~P. Kingma and Jimmy Ba.
\newblock Adam: {A} method for stochastic optimization.
\newblock In Yoshua Bengio and Yann LeCun, editors, \emph{3rd International
  Conference on Learning Representations, {ICLR} 2015, San Diego, CA, USA, May
  7-9, 2015, Conference Track Proceedings}, 2015.
\newblock URL \url{http://arxiv.org/abs/1412.6980}.

\bibitem[Ren et~al.(2015)Ren, He, Girshick, and Sun]{ren2015faster}
Shaoqing Ren, Kaiming He, Ross Girshick, and Jian Sun.
\newblock Faster r-cnn: Towards real-time object detection with region proposal
  networks.
\newblock \emph{Advances in neural information processing systems}, 28, 2015.

\end{thebibliography}

\newpage
\appendix

\section{Other NeSy Predictors}
\label{sec:other-approaches}

In this appendix, we outline the NeSy prediction approaches used in our experiments and then show that they share deterministic RSs as DPL.

The \textbf{semantic loss} (SL) \citep{xu2018semantic} is a penalty term that encourages a neural network to place all probability mass on predictions that are consistent with prior knowledge $\BK$.
In our setting, the SL is applied to a predictor $p_\theta(\vY \mid \vC)$ placed on top of a concept extractor $p_\theta(\vC \mid \vX)$ and it can be written as:
\[
    \textstyle
    \SL(p_\theta, (\vx, \vy), \BK) :=
            - \log \sum_\vc
            \Ind{ (\vc, \vy) \models \BK } p_\theta( \vc \mid \vx)
    \label{eq:semantic-loss}
\]

%
Like DPL, the SL relies on knowledge compilation to efficiently implement \cref{eq:semantic-loss}.
%
%
Importantly, if the distribution $p_\theta(\vc\mid \vx)$ allocates mass \textit{only} to concepts $\vc$ that satisfy the knowledge given $\vy$, these are optimal solutions, in the sense that the SL is exactly \textit{zero}.  We will make use of this fact in \cref{sec:det-opts-are-shared}.

During training, the SL is combined with any regular supervised loss $\ell$, for instance the cross-entropy, leading to the overall training loss:
\[
    \textstyle
    \frac{1}{|\dataset|} \sum_{(\vx, \vy) \in \dataset}
        \ell(p_\theta, (\vx, \vy)) + \mu \SL(p_\theta, (\vx, \vy), \BK)
    \label{eq:empirical-semantic-loss}
\]
with $\mu > 0$ a hyperparameter.  During inference, the SL plays no role:  the predicted label is obtained by simply taking the most likely configuration through a forward pass over the network.

\textbf{Logic tensor networks} (LTNs) \citep{donadello2017logic} are another state-of-the-art NeSy architecture that combines elements of reasoning- and penalty-based approaches.
The core idea behind LTNs, and of all other NeSy predictors based of fuzzy logic \citep{giannini2018convex}, is to \textit{relax} the prior knowledge $\BK$ into a real-valued function $\calT[\BK]$ \textit{quantifying} how close a prediction is to satisfying $\BK$.
LTNs perform this transformation using \textit{product real logic} \citep{donadello2017logic}.
In our context, this function takes the form $\calT[\BK]: [0, 1]^k \times [0, 1]^n \to [0, 1]$, and it takes as input the probabilities of the various concepts $\vC$ and labels $\vY$ and outputs a degree of satisfaction.

Crucially, fuzzy logics are designed such that, if all probability mass is allocated to configurations $(\vc, \vy)$ that \textit{do} satisfy the logic, then the degree of satisfaction is exactly $1$, \ie maximal.
We will leverage this fact in \cref{sec:det-opts-are-shared}.

During training, LTNs guide the concept extractor $p_\theta(\vC \mid \vX)$ towards predicting concepts that satisfy the prior knowledge $\BK$ by penalizing it proportionally to how far away their predictions are from satisfying $\BK$ given the ground-truth label $\vy$, that is, $1 - \calT[\BK](p(\vC \mid \vx), \Ind{\vY = \vy})$.
During inference, LTNs first predict the most likely concepts $\hat\vc = \argmax_{\vc} \ p_\theta(\vc \mid \vx)$ using a forward pass through the network, and then predict a label $\hat\vy$ that maximally satisfies the knowledge given $\hat\vc$, again according to $\calT[\BK]$.


\subsection{Deterministic Optima are Shared}
\label{sec:det-opts-are-shared}

The bulk of our theoretical analysis focuses on DPL because it offers a clear probabilistic framework, however in the following we show that deterministic optima do transfer to other NeSy predictors approaches.
Specifically, under \textbf{A2}, DPL, SL, and LTN all admit the same deterministic optima (det-opts).

To see this, let $\calC_{\vy}$ be the set of concepts vectors $\vc$ from which the label $\vy$ can be inferred, that is, $\calC_{\vy} = \{ \vc \in \calC \ : \ p^*(\vy \mid \vc) > 0 \}$.
Now, take a deterministic RS $p_\theta(\vC \mid \vX)$ for DPL, \ie a concept distribution that satisfies \cref{def:reasoning-shortcut} when the likelihood is computed as in \cref{eq:dpl-learning-objective}.
By determinism, this distribution can be equivalently written as a function mapping inputs $\vx$ to concept vectors $\hat\vc(\vx) \in \calC$, that is, $p_\theta(\vC \mid \vX = \vx) = \Ind{ \vC = \hat\vc(\vx) }$.
At the same time, by optimality and \textbf{A2} we have that, for every $(\vx, \vy) \in \dataset$, the NeSy predictor $p_\theta(\vY \mid \vX; \BK) = \sum_{\vc} 
\ u_{\BK}(\vY \mid \vc) \cdot p_\theta(\vc \mid \vX)$ allocates all probability mass to the correct label $\vy$.
As a consequence, $\hat\vc(\vx) \in \calC_{\vy}$, for every $(\vx, \vy) \in \dataset$.

Consider a NeSy predictor that first predicts concepts using $p_\theta(\vC \mid \vX)$ and then labels using a reasoning layer based on fuzzy logic, such that a prediction $\hat\vy \in \calY$ is chosen so that it minimial distance from satisfaction w.r.t. $\BK$, and the distance from satisfaction is also used as a training objective.
Since $\hat\vc \in \calC_{\vy}$ for every $(\vx, \vy) \in \calD$, by definition of T-norms, the label with minimal distance from satisfaction will necessarily be $\vy$: all other labels cannot be inferred from the knowledge, so they have larger distance from satisfaction.
Hence, training loss will be minimal for this predictor as well, meaning that $p_\theta(\vC \mid \vX)$ is a deterministic RS for it too.

%
Next, consider a NeSy penalty-based predictor where predictions are obtained by first inferring concepts using MAP over the concept extractor, that is, $\hat\vc = \argmax_{\vc} p_\theta(\vc \mid \vX)$, and then performing a forward pass over a neural prediction layer $p_\theta(\vY \mid \vC)$.
By construction, the penalty will be minimal whenever $(\vy, \hat\vc(\vx)) \models \BK$.  We already established that this is the case for all $(\vx, \vy) \in \dataset$, meaning that $p_\theta(\vC \mid \vX)$ is a deterministic RS for penalty-based predictors too.

\section{Proofs}
\label{sec:proofs}

In the proofs, we suppress $\BK$ from the notation for readability.

\subsection{Proof of \cref{lemma:abstraction-from-lh}: Upper Bound of the Log-Likelihood}

\textbf{Proof plan.}  The proof of the first point of our claim is split in three parts:
\begin{enumerate}[leftmargin=1.25em]

    \item We show that in expectation the log-likelihood in \cref{eq:dpl-learning-objective} is upper bounded by a term containing the KL divergence for $p_\theta (\vY \mid \vG)$. 

    \item We prove that, under \textbf{A1}, any optimum of \cref{eq:dpl-learning-objective} minimizes also the KL. In this step, we make use of Information Theory \citep{cover1999elements} to connect the two.

    \item We show that assuming \textbf{A1} and \textbf{A2} the optima $p_\theta(\vY \mid \vG)$ for the KL are given only by optima $p_\theta(\vY \mid \vX)$ for \cref{eq:dpl-learning-objective}.

\end{enumerate}
We proceed to prove the second point by leveraging the fact that $p_\theta(\vC \mid \vG)$ is given by marginalizing $p_\theta(\vC \mid \vX)$ over the generating distribution $p^*(\vX \mid \vG, \vS)$.

\textbf{Point (\textit{i}).}
(1) We upper bound the log-likelihood in \cref{eq:dpl-likelihood} as follows:
\begin{align}
    \calL(p_\theta, \dataset)
    & = \bbE_{\vg \sim p(\vG)} \bbE_{\vs \sim p(\vS)} \bbE_{\vx \sim p^*(\vX \mid \vg, \vs)} \bbE_{\vy \sim p^*(\vY \mid \vg)} \big[ \log p_\theta (\vy \mid \vx) \big]
    \\
    & = \bbE_{\vg \sim p(\vG)} \bbE_{\vx \sim p^*(\vX \mid \vg)} \bbE_{\vy \sim p^*(\vy \mid \vg)} \big( \log p_\theta(\vy \mid \vx) \big)
    \\
    & \leq \bbE_{\vg \sim p(\vG)} \bbE_{\vy \sim p^*(\vY \mid \vg)} \big( \log \bbE_{\vx \sim p^*(\vX \mid \vg) } [ p_\theta(\vy \mid \vx) ] \big)
    \\
    & = \bbE_{\vg \sim p(\vG)} \bbE_{\vy \sim p^*(\vY \mid \vg)} \big( \log p_\theta(\vy \mid \vg) \big)
    \\
    & = \bbE_{\vg \sim p(\vG)} \bbE_{\vy \sim p^*(\vY \mid \vg)} \Big( \log \frac{p_\theta(\vy \mid \vg)}{p^*(\vy \mid \vg)} - \log  p^*(\vy \mid \vg) \Big)
    \\
    & = \bbE_{\vg \sim p(\vG)} \big( - \KL[ p^*(\vy \mid \vg) \| p_\theta (\vy \mid \vg)] - \Ent [ p^*(\vy \mid \vg) ]  \big) 
    \label{eq:ll-le-kl-minus-h}
\end{align}
In the second line, we introduced the distribution $p^*(\vx \mid \vg) := \bbE_{\vs \sim p(\vS)} [p^*(\vx \mid \vg, \vs)] $, in the third line we applied Jensen's inequality, and in the fifth line we added and subtracted $p^*(\vy \mid \vg)$.
This proves our claim.

(2) We proceed showing that for a deterministic $p^*(\vX \mid \vG, \vS)$ every optimum $p_\theta(\vY \mid \vX)$ for the log-likelihood leads to an optimum $p_\theta(\vY \mid \vG)$ for the RHS of \cref{eq:optima-ineq}.  
We can rewrite the joint distribution as: 
\begin{align}
    \label{eq:posterior-distr-data}
    p(\vx, \vy)
        & := \bbE_{\vg \sim p(\vG)} \bbE_{\vs \sim p(\vS)} [ p^*(\vx \mid \vg, \vs) p^*(\vy \mid \vg) ]
    \\
        & = \bbE_{\vg \sim p(\vG)} \big[ p^*(\vy \mid \vg) \ \bbE_{ \vs \sim p(\vS) }[p^*(\vx \mid \vg, \vs)] \big]
    \\
        & = \bbE_{\vg \sim p(\vG)} \big[ p^*(\vy \mid \vg) p^*(\vx \mid \vg) \big]
    \\
        & = \bbE_{\vg \sim p(\vG)} \big[ p^*(\vy \mid \vg) p^*(\vg \mid \vx) p(\vx) / p(\vg) \big]
    \\
        & = \bbE_{\vg \sim p^*(\vG \mid \vx)}[ p^*(\vy \mid \vg) ] p(\vx)
    \\
        & = p^*(\vy \mid \vx) p(\vx)
\end{align}
where $p^*(\vy \mid \vx) = \bbE_{\vg \sim p^*(\vg \mid \vx)} [ p^*(\vy \mid \vg) ] $ and $p^*(\vG \mid \vx)$ is the posterior underlying the data generative process.
Hence, the log-likelihood in \cref{eq:dpl-likelihood} can be equivalently written as:
\begin{align}
    \bbE_{(\vx, \vy) \sim p(\vX, \vY)} [ \log p_\theta(\vy \mid \vx) ]
        &= \bbE_{\vx \sim p(\vX)} \bbE_{\vy \sim p^*(\vY \mid \vx)} [\log p_\theta (\vy \mid \vx)]
    \\
        & = \bbE_{\vx \sim p(\vX)} \bbE_{\vy \sim p^*(\vY \mid \vx)} \Big[ \log \frac{p_\theta (\vy \mid \vx)}{p^*(\vy \mid \vx)} - \log p^*(\vy \mid \vx) \Big]
    \\
        & = \bbE_{\vx \sim p(\vX)}\big( - \KL[ p^*(\vy \mid \vx) \| p_\theta(\vy \mid \vx) ]  - \Ent [p^*(\vy \mid \vx)] \big)
        \label{eq:ll-is-kl-minus-h}
\end{align}
In the first line, we used $p(\vX , \vY) = p^*(\vY \mid \vX) p(\vX)$, and then added and subtracted $\log p^*(\vy \mid \vx)$.
By comparing \cref{eq:ll-le-kl-minus-h} and \cref{eq:ll-is-kl-minus-h}, we obtain:
\begin{align}
    \label{eq:dpl-ineq-app}
    \textstyle
        & \bbE_{\vx \sim p(\vX)} \big(
            - \KL [ p^*(\vY \mid \vx) \| p_\theta(\vY \mid \vx) ] - \Ent [ p^*(\vY \mid \vx) ]
        \big)
        \nonumber
        \\
        & \qquad \leq
        \bbE_{\vg \sim p(\vG)} \big(
            - \KL [ p^*(\vY \mid \vg) \| p_\theta(\vy \mid \vg) ] - \Ent [ p^*(\vY \mid \vg) ]
        \big)
\end{align}
Now, consider a distribution $p_\theta(\vY \mid \vX)$ that attains maximal likelihood.  Then, $\KL[ p^*(\vy \mid \vx) \| p_\theta(\vy \mid \vx) ] = 0$, and we can rearrange the inequality in \cref{eq:dpl-ineq-app} to obtain:
\[
    \label{eq:optima-ineq}
    \bbE_{\vg \sim p(\vG)} \big( \KL [ p^*(\vy \mid \vg) \| p_\theta(\vy \mid \vg) ] \big) \leq \Ent [ \vY \mid \vX ] - \Ent [ \vY \mid \vG ]
\]
Here, $\Ent [ \vY \mid \vX ] = \bbE_{\vx \sim p(\vX)} [ \Ent[p^*(\vY \mid \vx) ]$ and $\Ent [ \vY \mid \vG ] = \bbE_{\vg \sim p(\vG)} \Ent[p^*(\vY \mid \vg)]$ are conditional entropies.

We want to show that, under \textbf{A1}, the right-hand side of \cref{eq:optima-ineq} is in fact zero.  
As for the other conditional entropy term, recall that \citep{cover1999elements}:
\[
    \Ent [\vY \mid \vX]
        = \Ent [ \vY \mid \vX ] - \Ent [\vY] + \Ent [\vY]
        = - \MI[ \vX : \vY ] + \Ent [ \vY ] 
\]
where $\MI[ \cdot, \cdot ]$ is the mutual information.
By the chain rule of the mutual information, cf. \citep[Theorem~2.8.1]{cover1999elements}, we have:
\[
    \MI [ \vY : \vX, \vG  ]
        = \MI [ \vY : \vX ] + \MI [ \vY : \vG \mid \vX ]
        = \MI [ \vY : \vG ] + \MI [ \vX : \vY \mid \vG ]
    \label{eq:mis}
\]
where $ \MI [ \cdot, \cdot \mid \cdot] $ is the conditional mutual information.
The structure of our generative process in \cref{fig:second-page} implies that $\MI [\vX : \vY \mid \vG] = 0$, so \cref{eq:mis} boils down to:
\[ 
    \MI [\vX : \vY] = \MI [ \vY : \vG ] - \MI [ \vY : \vG \mid \vX ]
    \label{eq:whatever-1}
\]
By \textbf{A1}, we have that:
\begin{align}
    \MI [ \vY : \vG \mid \vX ] &= \Ent [ \vG \mid \vX ] - \Ent [ \vG \mid \vX, \vY ] \\
    &= \Ent [ f^{-1}_{1:k}(\vX) \mid \vX ] - \Ent [ f^{-1}_{1:k}(\vX) \mid \vX, \vY ] \\ 
    &= \Ent [ f^{-1}_{1:k}(\vX) \mid \vX ] - \Ent [ f^{-1}_{1:k}(\vX) \mid \vX ] = 0
\end{align}
Plugging this into \cref{eq:whatever-1} entails that $\MI [ \vX : \vY ] = \MI [\vY : \vG] $, or equivalently that $ \Ent [\vY : \vX] = \Ent [ \vY : \vG ]$.
This means that the right-hand side of \cref{eq:optima-ineq} is indeed zero, which entails that the $\KL$ is zero and that therefore $p_\theta(\vY \mid \vG)$ optimizes the right-hand side of \cref{eq:dpl-upper-bound}.

(3) We proceed showing that by assuming also \textbf{A2} whatever optimum $p_\theta(\vY \mid \vG)$ is identified by $p_\theta(\vY \mid \vX)$ that is also optimum.
First, note that under \textbf{A2} $p^*(\vY \mid \vG)$ is deterministic, so we have $\Ent [\vY \mid \vG] = 0$. With \textbf{A1} and \textbf{A2}, both $p^*(\vy \mid \vg)$ and $p^*(\vy \mid \vx)$ are deterministic and therefore \cref{eq:dpl-ineq-app} can be rewritten as:
\[
    \bbE_{\vx \sim p(\vX)} [
        \log p_\theta(\vY = (\beta_\BK \circ f^{-1}_{1:k})(\vx) \mid \vx)
    ]
    \leq
    \bbE_{\vg \sim p^*(\vG)}
        [\log p_\theta(\vY = \beta_\BK(\vg) \mid \vg)] 
\]
In particular, for each $\vx$ the maximum of the log-likelihood is $0$ and it is attained when the label probability is one.  We make use of this observation in the next step.

Next, we show that the bound in \cref{eq:dpl-ineq-app} is in fact \textit{tight}, in the sense that whenever $p_\theta ( \vY \mid \vG ) $ maximizes the left-hand side, $p_\theta(\vY \mid \vX)$ maximizes the right-hand side.
We proceed by contradiction. 
Fix $\vg$ and let $\calO$ be the set of those $\vs$ for which $p_\theta(\vy \mid \vx = f(\vg, \vs) ) < 1$ and assume that it has non-vanishing measure.
Then, the posterior distribution is also strictly less than one:
\begin{align}
    p_\theta(\vy \mid \vg)
        & = \bbE_{\vs \sim p(\vS)} \bbE_{\vx \sim p^*(\vX \mid \vg, \vs)} [ p_\theta(\vy \mid \vx) ]
    \\
        & = \int_{\bbR^q} p(\vs)  \int_{\bbR^d} p_\theta (\vy \mid \vx) p^*(\vx \mid \vg, \vs) \mathrm d \vx \mathrm d \vs
    \\
        & = \int_{\bbR^s } p(\vs) \int_{\bbR^d} \delta \{ \vx - f(\vg, \vs) \} \mathrm d \vx \mathrm d \vs - \int_{\bbR^s} p(\vs) \int_{\bbR^d} [1 - p_\theta (\vy \mid \vx)] \delta \{ \vx - f(\vg, \vs) \} \mathrm d \vx \mathrm d\vs
    \\
        & = 1 - \int_{\calO} (1 - p_\theta (\vy \mid \vx = f(\vg, \vs))) p(\vs) \mathrm d \vs
    \\
        & < 1
\end{align}
%
Therefore, there cannot be any optimal solutions $p_\theta(\vY \mid \vG)$ that are given by non-optimal probabilities $p_\theta (\vY \mid \vX)$.
This proves the claim.

\textbf{Point (\textit{ii}).} 
%
%
We consider now which distributions $p_\theta(\vC \mid \vG)$ correspond to a unique distribution $p_\theta(\vC \mid \vX)$. First, we define as $\calP$ the set of candidate distributions $p_\varphi (\vC \mid \vX)$, with $\varphi \in \Theta$, for which it holds:
\[
    \bbE_{\vs \sim p^*(\vS)}\bbE_{\vx \sim p^*(\vX \mid \vg, \vs)}[ p_\varphi(\vC \mid \vx) ] = p_\theta( \vC \mid \vg)
\]
For all distributions of the form $p_\theta(\vC \mid \vG) = \Ind{\vC = \vc} $, for $\vc \in \calC$, $\calP$ restricts to a single element, i.e., $p_\varphi(\vC \mid \vx) = \Ind{\vC = \vc}$. 
We proceed by contradiction and consider a set $\calO_\vX$ of non-vanishing measure such that:
\[
    p_\varphi(\vC \mid \vx) \neq \Ind{\vC = \vc}, \quad \forall \vx \in \calO_\vX
\]
Let $p^*(\vX\ \mid \vG) = \bbE_{\vs \sim p^*(\vS)} p^*(\vX \mid \vG, \vS)$.  Then, it holds that:
\begin{align} \label{eq:contazzo}
    p_\varphi(\vC \mid \vG) &= \int_{ \calX } p_\varphi(\vC \mid \vx) p^*(\vx \mid \vG) \mathrm d \vx \nonumber \\
        &= \int_{ \calX \setminus \calO_\vX } \Ind{ \vC = \vc } p^*(\vx \mid \vG) \mathrm d \vx + \int_{\calO_\vX} p_\varphi(\vC \mid \vx) p^*(\vx \mid \vG) \mathrm d \vx \nonumber \\ 
        &= (1 - \lambda) \cdot \Ind{ \vC = \vc } + \lambda \cdot \Tilde p_\varphi(\vC \mid \vG)
\end{align}
where we denoted $ \Tilde p_\varphi (\vC \mid \vG)$ the normalized probability distribution given by integrating $p_\varphi(\vC \mid \vX)$ on $\calO_\vX$ solely, and $\lambda$ is the measure of $\calO_\vX$. 
Notice that the RHS of \cref{eq:contazzo} is exactly $\Ind{\vC = \vc}$ \textit{iff} $\lambda = 0$ or $ \Tilde p_\varphi (\vC \mid \vG) = \Ind{\vC = \vc}$, which contradicts the claim.
Hence, all probabilities $p_\theta(\vC \mid \vG) = \Ind{\vC = \vc} $ are only given by probabilities $p_\theta(\vC \mid \vx) = \Ind{\vC = \vc} $, for all $\vx \in \calX$. This yields the claim.

\subsection{Proof of \cref{thm:mc-det-opts}: Counting the Deterministic Optima}
\label{sec:proof-of-mc-det-opts}

We want to count the number of deterministic optima of the log-likelihood in \cref{eq:dpl-likelihood}.
Recall that, by \cref{lemma:abstraction-from-lh}, under \textbf{A1} and \textbf{A2} any optimum $p_\theta(\vY \mid \vX)$ of \cref{eq:dpl-learning-objective} yields an optimum $p_\theta(\vY \mid \vG)$ of the upper bound in \cref{eq:dpl-upper-bound}.
Following, by point (\textit{ii}) of \cref{lemma:abstraction-from-lh} we have that deterministic optima are shared between $p_\theta(\vC \mid \vX)$ and $p_\theta(\vC \mid \vG)$.
This means that we can equivalently count the number of deterministic optima $p_\theta(\vC \mid \vG)$ for the upper bound in \cref{eq:dpl-upper-bound}.  We proceed to do exactly this.

Let $\calA$ be the set of all possible maps $\alpha: \vg \mapsto \vc$, each inducing a candidate concept distribution $p_\theta(\vC \mid \vG) = \Ind{\vC = \alpha(\vG)}$.
The only $\alpha$'s that achieve maximal likelihood are those that satisfy the knowledge for all $\vg \in \mathsf{supp}(\vG)$ for the learning problem, that is: 
\[
    \beta_\BK (\vg) = (\beta_\BK \circ \alpha) (\vg),
\]
\ie that it is indeed the case that the concepts output by $\alpha(\vg)$ predict the ground-truth label $h_\BK(\vg)$. 
%
Notice that, only one of them is correct and coincides with the identity, \ie $\alpha(\vg) = \vg$.
These $\alpha$ are those that satisfy the knowledge on all examples, or equivalently the \textit{conjunction} of the knowledge applied to all examples, that is:
\[
    \bigwedge_{\vg \in \dataset_\vG} \big( (\beta_\BK \circ \alpha)(\vg) = h_\BK(\vg) \big)
\]
This means that only a subset of $\calA$ contains those maps consistent with the knowledge. The total number of these maps is then given by:
\[
    \textstyle
    \sum_{\alpha \in \calA} \Ind{
        \bigwedge_{\vg \in \dataset_{\vG}}
            (\beta_\BK \circ \alpha)(\vg) = \beta_\BK(\vg) \big)
    }
    \label{eq:model-count-app}
\]
This yields the claim.

\subsection{Proof of \cref{prop:structure-of-nondet-ops}: Link Between Deterministic and Non-deterministic Optima}

\textbf{Point (\textit{i})}.
We begin by proving that, for DPL, any convex combination of optima of the likelihood \cref{eq:dpl-likelihood} is itself an optimum.
Fix any input $\vx$.
First, $p^*(\vy \mid \vx)$ is the optimal value of the log-likelihood in \cref{eq:dpl-learning-objective}, according to \cref{eq:optima-ineq} from point (\textit{i}) of \cref{lemma:abstraction-from-lh}.
Now, let $p^{(1)} (\vC \mid \vx) $ and $p^{(2)} (\vC \mid \vx)$ be two concept distributions that both attain optimal likelihood, \ie for $i \in \{1, 2\}$ it holds that $\sum_\vc u_\BK(\vy \mid \vc) \cdot p^{(i)} (\vc \mid \vx) = p^*(\vy \mid \vx)$.
The likelihood term of any convex combination of the two optima is given by:
\begin{align}
    & \sum_\vc u_\BK(\vy \mid \vc) [
            \lambda p^{(1)} (\vc \mid \vx) + (1 - \lambda) p^{(2)} (\vc \mid \vx )
        ]
    \\
        & = \lambda \sum_\vc
            u_\BK(\vy \mid \vc) p^{(1)}(\vc \mid \vx)
        + (1 - \lambda) \sum_\vc
            u_\BK(\vy \mid \vc) p^{(2)}(\vc \mid \vx)
    \\
        & = \lambda p^*(\vy \mid \vx) + (1 - \lambda) p^*(\vy \mid \vx)
    \\
        & =  p^*(\vy \mid \vx)
\end{align}
Hence, the convex combination is also an optimum of the likelihood.
Note that the very same reasoning applies to the Semantic Loss (\cref{eq:semantic-loss}), again due to linearity of the expectation over $\vC$.

\begin{remark}
    \changed{Since SL and DPL limit $p_\theta (\vC \mid \vX)$ to be factorized as $\prod_i p(C_i \mid \vX)$, some convex combinations that would be in principle solutions as per point \textbf{(i)} cannot be expressed. Essentially, this translates into an additional constraint for the solutions of SL and DPL that is, given two factorized probabilities $p^{(1)} (\vC \mid \vx) $ and $p^{(2)} (\vC \mid \vx) $, then: 
    \[
        p(\vC \mid \vx) = \lambda p^{(1)} (\vC \mid \vx) + (1 - \lambda) p^{(2)} (\vC \mid \vx) \text{ is a solution} \iff p(\vC \mid \vx) \text{ is factorized } \forall x
    \]
    This applies only to models that integrate probabilistic logic by predicting the concepts independently, whereby relaxations of SL and DPL, like \citep{ahmed2022semantic, AhmedKLR23}, could naturally express arbitrary convex combinations of deterministic RSs.
    }
\end{remark}

%
%
\begin{remark}
The above result does not hold for LTNs, in general.  We show this by constructing a counter-example
for the default choice of T-conorms ~\citep{donadello2017logic}.
That is, a convex combination of deterministic optima that is \textit{not} itself an optimum.
Recall that LTN uses \textit{product real logic} to define a degree of knowledge satisfaction and consider the prior knowledge $\BK = (C_1 \lor C_2)$, where $C_1$ and $C_2$ are two distinct binary concepts.  In LTN, the degree of satisfaction of $\BK$ is given by the T-conorm of the logical disjunction, which is defined as $S(a, b) = a + b - a b$.
Fix an input $\vx$ and consider two deterministic distributions $p^{(1)}(\vC \mid \vx) = \Ind{\vC = (1, 0)^\top}$ and $p^{(2)}(\vC \mid \vx) = \Ind{\vC = (0, 1)^\top}$.  It is clear that both distributions satisfy the prior knowledge and as such are optimal.
Now take any convex combination $p (\vC) = \lambda p^{(1)} (\vC)  + (1-\lambda) p^{(2)} (\vC)$.  Then, $a = \lambda p^{(1)}(C_1 = 1) + (1-\lambda) p^{(2)}(C_1 = 1) = \lambda$ and, for similar reasons, $b = 1 -\lambda$.
Then, it holds:
\begin{align}
    S(p(C_1=1), p(C_2=1))
        & = \lambda + (1-\lambda) - \lambda(1 -\lambda) \\
        & = 1 - \lambda + \lambda^2
        \leq 1
\end{align}
where the equality holds \textit{iff} $\lambda \in \{ 0, 1 \}$.
\end{remark}

\textbf{Point (\textit{ii}).}  Under \textbf{A1} and \textbf{A2}, we can count all deterministic solutions $p_\theta(\vC \mid \vG)$ via \cref{thm:mc-det-opts}.  Here, we show that these deterministic optima constitute a \textit{complete} basis for all optimal solutions of \cref{eq:optima-ineq}.
From (\textit{i}), we have that any convex combination of the deterministic optima is also an optimum. 
We will show that these are the \textit{only} optimal solutions. 

Recall that the optimal solutions are all of the form $p^*(\vY = \beta_\BK(\vg) \mid \vg) = 1$, as a consequence of \textbf{A1} and \textbf{A2} from \cref{lemma:abstraction-from-lh}, for all $\vg \in \mathsf{supp}(\vG)$.
Notice that any optimal solution $p_\theta(\vC \mid \vg)$ must place mass to those concepts that lead to the correct label. Formally, for every $\vg$ and $\vy = \beta_\BK(\vg)$, it holds:
\begin{align}
    p_\theta(\vy \mid \vg)
        &= \sum_\vc u_\BK (\vy \mid \vc) p_\theta (\vc \mid \vx)
    \\
        & \leq \sum_{\vc} p_\theta(\vc \mid \vg) = 1
\end{align}
where the equality holds \textit{iff} $u_\BK(\vy \mid  \vc) = 1$ for all $p_\theta(\vc \mid  \vx) > 0 $. In other words, this means that each $\vc \sim p_\theta( \vC \mid \vg) $ must be an optimal solution for the logic.
This proves the claim.

%

\textbf{Point (\textit{iii})}.
If \textbf{A2} does not hold, there can be optima solutions that are given as convex combinations of non-optimal deterministic probabilities.
First, notice that according to \cref{lemma:abstraction-from-lh}, point (\textbf{i}), the optimal solutions for $p_\theta (\vY \mid \vX; \BK)$ minimize the $\KL$ term and are equivalent to $p^*(\vY \mid \vX)$.  
Then, consider for a given $\vx$ an optimal solution $p(\vC \mid \vx) = \lambda \Ind{ \vC = \vc_1 } + (1- \lambda) \Ind{\vC = \vc_2}$ that is a convex combination of two deterministic probabilities. From the convexity of the $\KL$ it holds:
\begin{align}
    & \KL [ p^*(\vy \mid \vx) \mid \mid \lambda u_\BK(\vy \mid \vc_1; \BK) + (1 - \lambda) u_\BK(\vy \mid \vc_2; \BK) ] \\
    & \leq  \lambda \cdot \KL [ p^*(\vy \mid \vx) \mid \mid  u_\BK(\vy \mid \vc_1; \BK)] + (1 - \lambda) \cdot \KL [ p^*(\vy \mid \vx) \mid \mid  u_\BK(\vy \mid \vc_2; \BK)]
\end{align}
where the equality holds \textit{iff} $u_\BK (\vy \mid \vc_1; \BK) = u_\BK (\vy \mid \vc_2; \BK) = p^*(\vy \mid \vx)$. This shows that \textit{there can exist solutions that are  convex combinations of non-optimal deterministic probabilities}. 
We proceed to show how the space of optimal solutions is defined.
On the converse, if $u_\BK (\vy \mid \vc_1; \BK) \Ind{\vC -\vc_1} $ is a solution, also $u_\BK (\vy \mid \vc_2; \BK) \Ind{\vC -\vc_2}$ must be a solution:
\begin{align}
    \sum_\vc u_\BK (\vy \mid \vc) [\lambda \Ind{\vC = \vc_1} + (1-\lambda) \Ind{\vC = \vc_2} ] &= p^*(\vy \mid \vx) \\
    \lambda u_\BK(\vy \mid \vc_1) + (1 - \lambda) u_\BK(\vy \mid \vc_2) &= p^*(\vy \mid \vx) \\
    \lambda p^*(\vy \mid \vx) + (1 - \lambda) u_\BK(\vy \mid \vc_2) &= p^*(\vy \mid \vx) \\
    (1 - \lambda) u_\BK(\vy \mid \vc_2) &= (1 -\lambda) p(\vy \mid \vx) \\
    u_\BK(\vy \mid \vc_2) &= p^*(\vy \mid \vx)
\end{align}

We now show that any optimum given by a generic $p_\varphi(\vC \mid \vx)$, with $\varphi \in \Theta$, can be expressed as a convex combination of (1) an optimum that is a convex combination of deterministic optima and (2) an optimum that is a convex combination of deterministic, but non-optimal, probabilities:
\begin{align}
    p^*(\vy \mid \vx) &= \sum_\vc u_\BK(\vy \mid \vc) p(\vc \mid \vx) \\
    &= \sum_{\vc \in \calC_\vy} u_\BK(\vy \mid \vc) p(\vc \mid \vx)  + \sum_{\vc \in \calS^c_{\vy}} u_\BK(\vy \mid \vc) p(\vc \mid \vx) \\
    &= \lambda \sum_{\vc \in \calC_\vy} u_\BK(\vy \mid \vc) \Tilde p(\vc \mid \vx)  + (1 - \lambda) \sum_{\vc \in \calS^c_{ \vy}} u_\BK(\vy \mid \vc) \bar p(\vc \mid \vx) 
\end{align} 
where $\calC_\vy = \{ \vc \in \calC: u_\BK = p^*(\vy \mid \vx) \} $ and $\calS^c_{\vy} = \calC / \calC_\vy$ are two disjoint sets. 
In the second line, we rewrote the summation on the two terms considering the two sets, whereas in the third line we introduced: $\lambda = \sum_{\vc \in \calC_\vy} p(\vc \mid \vx)$ and $\Tilde p(\vc \mid \vx) = p(\vc \mid \vx) / \lambda $, $1 - \lambda = \sum_{\vc \in \calS_{\bar \vy}} p(\vc \mid \vx)$ and $\bar p(\vc \mid \vx) = p(\vc \mid \vx) / (1 - \lambda)$. 
Since each $\Tilde{p}(\vc \mid \vx) $ lead to an optimum by construction, it must be that also $\Bar{p}(\vc \mid \vx) $ is an optimum, by the previous point.
In general, there can be many $\Bar{p}(\vc \mid \vx)$ leading to optima, even when $\Tilde{p}(\vc \mid \vx)$ reduces to only the ground-truth element, \ie $\Tilde{p}(\vc \mid \vx) = \Ind{ \vC - f^{-1}_{1:k}(\vx)}$. 
Notice that this must hold for all $\vx \sim p^*(\vX)$.
This proves the claim.

\subsection{Proof of \cref{prop:multitask}: Multi-task Learning}

When considering multiple task $T$, suppose that, for each $\vx \in \bigcap_t \dataset_t$, we get $T$ different labels $\vy^{(1)}, \ldots, \vy^{(T)}$, each given in accordance to knowledge $ \BK^{(1)}, \ldots, \BK^{(T)}$, respectively. Similarly to \cref{eq:dpl-learning-objective}, we consider the learning objective for all tasks with the joint log-likelihood term:
\[
    \log \prod_{t=1}^T p_\theta (\vy^{(t)} \mid \vx; \BK^{(t)})  
    = \sum_{t=1}^T  \log p_\theta (\vy^{(t)} \mid \vx; \BK^{(t)}) 
\]
Under \textbf{A1} and \textbf{A2}, \cref{lemma:abstraction-from-lh} point (\textit{i}) holds for each task $t$. In the following, we denote with $\beta_\BK^{(t)}$ the underlying maps for each $\BK^{(t)}$, which by \textbf{A2} are deterministic. Hence, the learning objective becomes:
\[
   \calL = \sum_{t=1}^T \bbE_{\vx \sim p(\vX)} [ \log p_\theta (\vY^{(t)} = (\beta_\BK^{(t)} \circ f^{-1}_{1:k}) (\vx) \mid \vx; \BK^{(t)})  ] 
\]
and, similarly to point (\textit{i}) of \cref{lemma:abstraction-from-lh}, we get the following upper-bound:
\[ \label{eq:upper-bound-multitask}
    \calL \leq  \sum_{t=1}^T \bbE_{\vg \sim p(\vG)} [ \log p_\theta (\vY^{(t)} = \beta_\BK^{(t)}  (\vg) \mid \vg; \BK^{(t)})  ] 
\]
Notice that the optimal values for \cref{eq:upper-bound-multitask} are given by those distributions that are one for each term in the summation. 
Then, following from \cref{thm:mc-det-opts}, we have that the deterministic maps $\alpha$'s that optimize \cref{eq:upper-bound-multitask} must be consistent with each task $t$ and for each $\vg \in \mathsf{supp}(\vG)$ satisfy:
\[  \label{eq:condition-of-mtl}
    \textstyle
    \bigwedge_{t=1}^T \big( ( \beta_\BK^{(t)} \circ \alpha )(\vg) = \beta_\BK^{(t)}(\vg) \big)
\]
which, equivalently to \cref{thm:mc-det-opts}, points to the condition that the maps $\alpha$ must be consistent with solving all tasks $t$. This exactly amounts to solving the conjunction of all knowledge $\BK = \bigwedge_t \BK^{(t)}$. 
This yields the claim.


\subsection{Proof of \cref{prop:concept-supervision}: Concept Supervision}

Let $p^*(\vG \mid \calS)$ be the distribution of ground-truth concepts restricted to a subset $\calS \subseteq \mathsf{supp}(\vG)$:
\[
    p^*(\vG \mid \calS) = \frac{1}{\calZ}  p^*(\vg) \Ind{\vg \in \calS}  \quad \mathrm{with} \; \calZ = \sum_\vg p^*(\vg) \Ind{\vg \in \calS}  
\]
and $p^*(\vX \mid \calS) = \bbE_{(\vg, \vs) \sim p^*(\vG \mid \calS) p^*(\vS)} p^*(\vX \mid \vg, \vs)$ be the corresponding restricted input distribution. 

Under \textbf{A1}, the expectation of the log-likelihood term for concept supervision in \cref{sec:mitigation-strategies} is:
\begin{align} \label{eq:upperbound-csup}
    \bbE_{\vx \sim p^*(\vX \mid \calS)} \log p_\theta(\vC_I = f^{-1}_I(\vx) \mid \vx) &= \bbE_{\vg \sim p^*(\vG \mid \calS)} \bbE_{\vs \sim p^*(\vS)} \log p_\theta ( \vC_I = \vg_I \mid f(\vg, \vs) ) \nonumber \\
    &\leq \bbE_{\vg \sim p^*(\vG \mid \calS)}  \log \bbE_{\vs \sim p^*(\vS)}[ p_\theta ( \vC_I = \vg_I \mid f(\vg, \vs) )] \nonumber \\
    &= \bbE_{\vg \sim p^*(\vG \mid \calS)} \log  p_\theta ( \vC_I = \vg_I \mid \vg )
\end{align}
where $p^*(\vX \mid \calS) =  \bbE_{\vg \sim p^*(\vG \mid \calS)} \bbE_{\vs \sim p^*(\vS)} p_\theta(\vX \mid \vg, \vs) $.
In the first line we made use of \textbf{A1} to write $\vx = f(\vg, \vs)$, in the second line we used Jensen's inequality, and then we introduced the marginal distribution $p_\theta(\vC \mid \vG) = \bbE_{\vs \sim p^*(\vs)}[ p_\theta(\vC \mid f(\vG, \vs))]$.
Recall that $I$ denotes the subset of the supervised ground-truth concepts and that the log-likelihood is evaluated only w.r.t. those concepts. 

Now notice that, given \textbf{A1}, the maximum for the LHS of \cref{eq:upperbound-csup} is zero, \ie coincides with maximum log-likelihood. Consistently the RHS of \cref{eq:upperbound-csup} is zero only when $p_\theta(\vC_I \mid \vg)$ places all mass on $\vg_I$.  We proceed considering those maps $\alpha: \vg \mapsto \vc$ which lead to deterministic, optimal solutions for the RHS.  For these, it must hold that for each $\vg \in \calS$:
\[  \textstyle
    \bigwedge_{i \in I} \alpha_i(\vg) = g_i
\]
When taken all together, we can estimate how many $\alpha \in \calA$ (cf. \cref{sec:proof-of-mc-det-opts}) satisfy the above condition for all $\vg$: 
\[  
    \textstyle
    \sum_{\alpha \in \calA} \Ind{ \bigwedge_{\vg \in \calS} \bigwedge_{i \in I} \alpha_i(\vg) = g_i }
\]
This yields the claim.

\subsection{Proof of \cref{prop:recon-loss}: Reconstruction}

Under \textbf{A1}, the reconstruction penalty can be written as:
\begin{align}
    \bbE_{\vx \sim p^*(\vX)} [\calR(\vx)] &= -\bbE_{(\vg,\vs) \sim p^*(\vG)p^*(\vS)} \bbE_{\vx \sim p^*(\vX \mid \vg, \vs)} \big[ \bbE_{(\vc,\vz) \sim p_\theta(\vC, \vZ \mid \vx) }  \log p_\psi(\vx \mid \vc, \vz)  \big]   \\
    &= - \bbE_{(\vg, \vs) \sim p^*(\vG)p^*(\vS) } \big[ \bbE_{(\vc, \vz) \sim p_\theta(\vC, \vZ \mid f ( \vg, \vs) )  }[ \log p_\psi ( f (\vg, \vs) \mid \vc, \vz )  ] \big]   
\end{align}
From \textbf{A3}, we have that the both encoder and decoder distributions are factorized:
\begin{align}
    p_\theta( \vc, \vz \mid f(\vg, \vs) ) &= p_\theta(\vc \mid \vg) p_\theta (\vz \mid \vs) \\
    p_\psi( f(\vg, \vs) \mid \vc, \vz) &= p_\psi(\vg \mid \vc) p_\psi (\vs \mid \vz)
\end{align}
%
%
This yields:
\begin{align}
    \bbE_{\vx \sim p(\vX)} [\calR(\vx)]
        &= - \bbE_{(\vg, \vs) \sim p^*(\vG)p^*(\vS)} \big[ \bbE_{\vc \sim p_\theta(\vC |  \vg) }[ \log p_\psi ( \vg| \vc)  ] +   \bbE_{(\vz) \sim p_\theta(\vZ |  \vs) }[ \log p_\psi ( \vs| \vz)  ] \big] \nonumber  \\
        &= - \bbE_{\vg \sim p^*(\vG)}  \bbE_{\vc \sim p_\theta(\vC |  \vg) } \log p_\psi ( \vg| \vc) - \bbE_{\vs \sim p^*(\vS)}  \bbE_{\vz \sim p_\theta(\vZ |  \vs) } \log p_\psi ( \vs| \vz) \nonumber   \\
        &\geq - \bbE_{\vg \sim p^*(\vG)}  \bbE_{\vc \sim p_\theta(\vC |  \vg) } \log p_\psi ( \vg| \vc)
        \label{eq:lower-bound-rec}
\end{align}
where in the first row we separated the two logarithms and removed the expectations on $p_\theta(\vZ \mid \vs)$ and $p_\theta(\vC \mid \vg)$ for the terms $p_\psi(\vg \mid \vc)$ and $ p_\psi(\vs \mid \vz) $, respectively. In the third line, we discarded the term with the match of the reconstruction of the style, giving the lower bound. 

When $\calR$ goes to zero, the lower bound also goes to zero and the last term of \cref{eq:lower-bound-rec} is minimized. This happens whenever, for each $\vc \sim p_\theta(\vc \mid \vg)$, it holds that $p_\psi(\vg \mid \vc) = 1$. 
This condition also implies that if for two different $\vg'$ and $\vg''$ there exist at least one $\vc$ such that $p_\theta (\vc \mid \vg') \cdot p_\theta (\vc \mid \vg'') > 0$, then $ p_\psi(\vg' \mid \vc)$ and $p_\psi(\vg'' \mid \vc)$ cannot both be optimal.

We restrict now to a deterministic map $\alpha:\vg \mapsto \vc$ for $p_\theta(\vC \mid \vG)$ and describe the condition when optimal decoders are attained, \ie $p_\psi(\vg \mid \vc)=1$ for some \vc. By the previous argument, an optimal map $\alpha$ that leads to optimal decoders must always map ground-truth concepts $\vg$ to different concepts $\vc$, that is:
\[
    \alpha(\vg') \neq \alpha(\vg'') \qquad \forall \vg' \ne \vg''
\]
In particular, this condition must hold for all different $\vg \in \mathsf{supp}(\vG)$. The number of all solutions is then given by:
\[  \label{eq:model-count-rec-app}
    \textstyle
        \sum_{\alpha \in \calA} \Ind{
        \bigwedge_{\vg \in \mathsf{supp}_\vG}
            \bigwedge_{\vg' \in \mathsf{supp}_\vG : \vg' \ne \vg} 
                \alpha(\vg) \neq \alpha(\vg') }
\]
as claimed.

\section{Experimental Details and Further Results} \label{sec:app-implementations}

We report here all further details concerning the experiments in \cref{sec:experiments}.

\subsection{Implementation}

The code of the experiments builds on top of \texttt{nesy-cl} \citep{marconato2023neuro} and \texttt{CBM-AUC} \citep{sawada2022concept}.  All experiments are implemented with Python 3.8.16 and Pytorch \citep{paszke2019pytorch} and run over one A100 GPU. {The code for the experiments is available at \href{https://github.com/ema-marconato/reasoning-shortcuts}{github.com/reasoning-shortcuts}.}

The implementation of DPL is taken from \citep{marconato2023neuro} and follows exactly \cref{eq:dpl-likelihood}, where to each world a label is assigned accordingly to the prior knowledge $\BK$. 
We implemented SL following the original paper \citep{xu2018semantic}, with the only difference that the prediction of the labels $\vY$ is done on top of the logits of $p_\theta(\vC \mid \vx)$. This relaxes the conditional independence between labels and concepts while being more in line with the generative process we assumed. 
The implementation of LTN is adapted from \texttt{LTN-pytorch} \citep{LTNtorch} and is based on the satisfaction loss introduced in \cref{sec:other-approaches}.

\subsection{Data sets \& Count of the Reasoning Shortcuts}
\label{sec:datasets}

Here, we illustrate how to count explicitly the number of deterministic RSs using the equations in \cref{tab:mitigation-strategies}, restricting ourselves to the case where no disentanglement is in place, for simplicity.
Similarly to \cref{sec:properties}, we assume \textbf{A1} and \textbf{A2} and that $\mathsf{supp}(\vG) = \calC = \calG$, and count the total number of deterministic optima (or \textit{det-opt}s for short) under different mitigation strategies.
In the following, $\calC_\vy \subseteq \calC = \calG$ refers to the set of $\vc \in \calC$ or $\vg \in \calG$ that are mapped by $\BK$ to label $\vy \in \calY$.  Note that $\sum_\vy |\calC_\vy| = |\calG|$.

\textbf{Explicit count for the likelihood.}  When the concept extractor $p_\theta(\vC \mid \vX)$ is sufficiently expressive, the deterministic mappings $\alpha$ are essentially arbitrary.  Specifically, the set $\calA$ of these $\alpha$'s includes all functions from $\calG$ to $\calC$.  These functions can be explicitly enumerated by counting how many ways there are to map each input vector $\vg$ to an arbitrary vector $\vc$.

Here, we are interested in counting the number of $\alpha$'s that attain optimal likelihood.  Each such $\alpha$ has to ensure that each $\vg \in \calG$ is mapped to a $\vc \in \calC_{\beta_\BK(\vg)}$ that yields the correct label, or in short, $\forall \vg \ . \ \alpha(\vg) \in \calC_{\beta_\BK(\vg)}$.
This condition is satisfied if and only if, for every $\vy$ in the data set, every $\vg \in \calC_{\vy}$ is mapped to a $\vc$ that is also in $\calC_{\vy}$, and there are exactly $|\calC_\vy|^{|\calC_\vy|}$ ways to do this.
This immediately shows that the overall number of \textit{det-opt}s $\alpha$ is:
\[  
    \textstyle
    \sum_{\alpha \in \calA} \Ind{ \bigwedge_{\vg \in \calG}  (\beta \circ \alpha)(\vg) = \alpha (\vg)  } = \prod_{\vy \in \calY}|\calC_\vy|^{|\calC_\vy|}
\]
This associates an explicit number to \cref{thm:mc-det-opts}.

\textbf{Explicit count for the reconstruction.}  The effect of adding a reconstruction penalty is that now, in order to achieve optimal loss, $\alpha$ has to map different $\vg$'s to distinct concepts $\vc$'s.  Unless this is the case, it becomes impossible to reconstruct the ground-truth concepts from the learned ones, and therefore the input $\vx$ generated from them.

The resulting computation is the same as above, except that now we have to associate the different $\vg$'s to different $\vc$'s \textit{without replacement}.  This yields the following count:
\[  
    \textstyle
    \sum_{\alpha \in \calA} \Ind{ \bigwedge_{\vg \in \calG}  (\beta \circ \alpha)(\vg) = \alpha (\vg)  } \cdot \Ind{ \bigwedge_{\vg, \vg' \in \calG: \vg \neq \vg'} \alpha(\vg) \neq \alpha (\vg')  } = \prod_{\vy \in \calY}|\calC_\vy|!
\]

\textbf{Explicit count for concept supervision.}  For the combination of logic and concept supervision, we suppose that for each equivalence class $\calC_\vy$, there are $\nu_\vy$ ground-truth concepts provided with supervision and that all concept dimensions $C_i$ receive this supervision. Let $\calS \subset \calG$ be the set of supervised concepts, such that $|\calS| = \sum_{\vy \in \calY} \nu_\vy$. 
From above, this means that we are specifying a total of $|\calS|$ ground-truth concepts.
From \cref{tab:mitigation-strategies}, we have that the number of \textit{det-opt}s amounts to:
\[  
    \textstyle
    \sum_{\alpha \in \calA} \Ind{ \bigwedge_{\vg \in \calG}  (\beta \circ \alpha)(\vg) = \alpha (\vg)  } \cdot \Ind{ \bigwedge_{\vg \in \calS} \bigwedge_{i=1}^k  \alpha_i(\vg) = g_i
    } = \prod_{\vy \in \calY} |\calC_\vy |^{ |\calC_\vy - \nu_\vy|}
\]
Finally, combining together the terms of prediction, reconstruction, and concept supervision, we obtain:
\begin{align}
    \textstyle
    \sum_{\alpha \in \calA} & \Ind{ \bigwedge_{\vg \in \calG}  (\beta \circ \alpha)(\vg) = \alpha (\vg)  } \times \\
    & \times \Ind{ \bigwedge_{\vg, \vg' \in \calG: \vg \neq \vg'}   \alpha(\vg) \neq \alpha (\vg')  } \times \\ 
    & \quad \times \Ind{ \bigwedge_{\vg \in \calS} \bigwedge_{i=1}^k  \alpha_i(\vg) = g_i
    } = \prod_{\vy \in \calY} |\calC_\vy  - \nu_\vy |!
\end{align}

\textbf{The disentangled case.}  If the network is \textit{disentangled}, the enumeration procedure becomes substantially more complicated and cannot be written compactly in closed form.  While the number of deterministic optima $\alpha$ can still be computed exactly using model counting \citep{darwiche2002knowledge, vergari2021compositional}, doing so is not necessary for the scope of our paper and therefore left to future work.

\subsubsection{Dataset: \XOR}

This dataset, introduced in \cref{ex:xor}, is a toy data set containing 3 bits $\vg = (g_1, g_2, g_3)$, for a total of $8$ possible combinations. The task consists in predicting the XOR operation among them, namely $ y = (g_1 \oplus g_2 \oplus g_3)$. The dataset is exhaustive and has no validation and test set. The model performances are evaluated on the training set. 

\textbf{Reasoning shortcut.}
For this dataset, we have $|\calC_0|=|\calC_1|=4$.
RSs arise depending on the structure of the underlying network.
When the ground-truth concepts are processed all together without any mitigation strategy, we obtain that the number of \textit{det-opt}s amounts to:
\[
    \prod_{y \in \calY} |\calC_y|^{|\calC_y|} = 4^4 \cdot 4^4
\]
The confusion matrices for all methods are reported in \cref{sec:confusion-matrices}.
On the other hand, we show that in the \textit{disentangled} case, only two combinations suffice to identify the correct solution, \eg
\[
    \begin{cases}
        \alpha(0) \oplus \alpha(0) \oplus \alpha(0) = 0 \\
        \alpha(0) \oplus \alpha(0) \oplus \alpha(1) = 1
    \end{cases}
\]
and here the only viable solution for the two is $\alpha(0) = 0$ and $\alpha(1)=1$. This condition is met in all our experiments in \textcolor{blue}{Table~2}.

\subsubsection{Dataset: \MNISTAdd}

We consider the version introduced in \citep{manhaeve2018deepproblog}, which consists of couples of digits, each ranging from $0$ to $9$, and the target consists in predicting the correct sum, \ie $y = g_1 + g_2$. This data set contains all possible combinations, for a total of $100$. The training set contains $42$k data, the validation set $12$k, and the test set $6$k.

\textbf{Reasoning shortcuts.}
RSs arise only as a result of the joint prediction of both digits. Notice that the number of elements $|\calC_y|$ for each sum $\vy$ can be evaluated as:
\[
    |\calC_y| = \begin{cases}
                    y + 1, \quad \mathrm{if} \; y \leq 9 \\
                    (18-y) + 1, \quad \mathrm{otherwise.}
                \end{cases}
\]
Therefore, the total number of \textit{det-opt}s amounts to:
\[
    \prod_{y \in \calY} |\calC_y|^{|\calC_y|} = {\prod_{y =1}^9 y^{2y}} \cdot  10^{10}
\]
When providing \textit{disentanglement}, the number of RSs reduce to $0$, as it sufficient to have the sums:
\[
    \alpha(c) + \alpha(c) = 2 \cdot c
\]
to uniquely identify the value of the digit $c$.

\subsubsection{Dataset: \MNISTShortcut}

This data set, proposed by \citep{marconato2023neuro}, is a biased version of \MNISTAdd, where only some combinations of the digits appear. We consider here a more challenging scenario w.r.t. the proposed version, consisting of the sums:
\[
    \begin{cases}
        \MZero + \MSix = 6 \\
        \MTwo + \MEight = 10 \\
        \MFour + \MSix  = 10 \\
        \MFour + \MEight = 12
    \end{cases}
    \quad \land \quad
    \begin{cases}
        \MOne + \MFive = 6 \\
        \MThree + \MSeven = 10 \\
        \MOne + \MNine  = 10 \\
        \MThree + \MNine = 12
    \end{cases}
\]
Overall, the training set contains $6720$ data, the validation set $1920$, and the test set $960$.

\textbf{Reasoning shortcuts.} We describe the RSs that arise even when the architecture incorporates \textit{disentanglement}. We evaluate the possible RSs empirically noticing that the system of observed sums can be written as a linear system, as done by Marconato et al. \citep{marconato2023neuro}:
\[
    \begin{cases}
        \alpha(0) + \alpha(6) = 6 \\
        \alpha(2) + \alpha(8) = 10 \\
        \alpha(4) + \alpha(6)  = 10 \\
        \alpha(4) + \alpha(8) = 12
    \end{cases}
    \quad \land \quad
    \begin{cases}
        \alpha(1) + \alpha(5) = 6 \\
        \alpha(3) + \alpha(7) = 10 \\
        \alpha(1) + \alpha(9)  = 10 \\
        \alpha(3) + \alpha(9) = 12
    \end{cases}
\]
Now, notice that we can find independent reasoning shortcuts for each of the two sides since they do not share any digits. For the LHS, we consider the sum $\alpha(2) + \alpha(8)=10$ and notice that we can find at most $10$ different attributions for having a correct sum. Notice that, some of them are not allowed as $\alpha(8)=0,1$ leads to inconsistent values for the fourth sum, and $\alpha(8)=3$ leads to an inconsistent sum for the first equation. So in total, for the LHS, we obtain $7$ possible solutions and, by symmetry, the same number also for the LHS. In total, the number of \textit{det-opt}s is equal to $7 \cdot 7$.

Experimentally, we consider limited concept supervision on $I=\{4,9\}$ which should be sufficient, in principle, to disambiguate between even and odds digits. This happens because specifying $\alpha(4)=4$ and $\alpha(9)=9$ admits only the ground-truth concepts as the optimal solution.

\subsubsection{Dataset: \MNISTOp}

In this dataset, we consider fewer combinations of digits, explicitly:
\[
    \begin{cases}
        \MZero + \MOne = 1 \\
        \MZero + \MTwo = 2 \\
        \MOne + \MThree = 4
    \end{cases}
\]
and similarly for multiplication. This data set contains $1680$ training examples, $480$ for the validation, and $240$ for the test set.

\textbf{Reasoning shortcuts.}
For the case of the addition task, we can have $2$ possible solutions, which are: 
\begin{itemize}
    \item $\alpha(0)=0$, $\alpha(1)=1$, $\alpha(2)=2$, and $\alpha(3)=3$;
    \item $\alpha(0)=1$, $\alpha(1)=0$, $\alpha(2)=1$, and $\alpha(3)=4$.
\end{itemize}
For multiplication, we have that since $\alpha(1) \cdot \alpha(3)=3$ it can be either $\alpha(1)=1$ or $\alpha(1)=3$. In both cases, it holds that $\alpha(0)=0$ and $\alpha(2)$ can be arbitrary. Hence, there are in total $2\cdot4$ possible \textit{det-opt}s.  
In \MTL, the reasoning shortcut for the addition does not hold since it leads to a sub-optimal solution for multiplication.

\subsubsection{Dataset: \BOIA}

This data set contains frames of driving scene videos for autonomous predictions \citep{xu2020boia}. Each frame is annotated with $4$ binary labels, indicating the possible actions, $\vY = (\texttt{move\_forward}, \texttt{stop}, \texttt{turn\_left}, \texttt{turn\_right})$. Each scene is also annotated with $21$ binary concepts $\vC$, underlying the \textit{reasons} for the possible actions, see \cref{tab:boia-concepts}.  The training set contains $16$k frames, with full label and concept supervision; the validation and the test set contain $2$k and $4.5$k annotated data, respectively. 

For designing the prior knowledge w.r.t. to the concepts in \cref{tab:boia-concepts}, we make use of the following rules for $\texttt{move\_forward}/ \texttt{stop}$ \textit{predictions}:
\[
\begin{cases}
    \texttt{red\_light} \implies \lnot \texttt{green\_light} \\
    \texttt{obstacle} = \texttt{car} \lor \texttt{person} \lor \texttt{rider} \lor \texttt{other\_obstacle} \\
    \texttt{road\_clear} \iff  \lnot \texttt{obstacle} \\
    \texttt{green\_light} \lor \texttt{follow} \lor \texttt{clear} \implies \texttt{move\_forward} \\
    \texttt{red\_light} \lor \texttt{stop\_sign} \lor \texttt{obstacle} \implies \texttt{stop}  \\\texttt{stop} \implies \lnot \texttt{move\_forward}  
\end{cases}
\]
For $\texttt{turn\_left}$, and similarly for $\texttt{turn\_right}$, we use:
\[
\begin{cases}
    \texttt{can\_turn} = \texttt{left\_lane} \lor \texttt{left\_green\_light} \lor \texttt{left\_follow} \\
    \texttt{cannot\_turn} = \texttt{no\_left\_lane} \lor \texttt{left\_obstacle} \lor \texttt{left\_solid\_line} \\
    \texttt{can\_turn} \land \lnot \texttt{cannot\_turn} \implies \texttt{turn\_left}
\end{cases}
\]

Notice that, since the concepts are predicted together, as explained in \cref{sec:app-architectures}, we can count the number of RSs as follows:
\begin{itemize}[leftmargin=1.25em]
    \item For $\texttt{move\_forward}$ and \texttt{stop}, the labels are predicted with the constraints such that $(\texttt{move\_forward}, \texttt{stop} )=(1,1)$ has no support. Hence, we consider only the predictions $(0,0), (0,1), (1,0)$. Next, we identify:
    \begin{itemize}
        \item $|\calC_{0,0}| = 1$, since it corresponds just to the case where no concepts are predicted.
        \item $|\calC_{(1,0)}| = 2^3 - 1$, which is the number of different concepts attribution for forward yielding a positive label;
        \item $ |\calC_{(0,1)}| = 280$ are the combination of the remaining concepts that yield the \texttt{stop} action, in agreement with the constraints. These were counted explicitly from the logic implementation.
    \end{itemize}
    Overall, the number of \textit{det-opt}s amount to $1 \cdot 7^7 \cdot 280^{280}$. 
    \item For $\texttt{turn\_left}$ and $\texttt{turn\_right}$, we count the cardinality of positive and negative predictions of the two classes $|\calC_0|$ and $|\calC_1|$:
    \begin{itemize}
        \item $|\calC_1| = 2^3 -1$, that are the only concepts attributions for the positive label;
        \item $|\calC_0|= 2^6 - |\calC_1|$, are all the remaining concept combinations.
    \end{itemize}
    For left and right, separately, we obtain that possible optimal solutions amount to $7^7 \cdot 57^{57}$.
\end{itemize}

Altogether, the count of \textit{det-opt}s for \BOIA goes as follows:
\[
    \prod_{\vy \in \calY} |\calC_\vy|^{|\calC_\vy|} = 1^1 \cdot 7^7 \cdot 280^{280} \cdot 7^7 \cdot 57^{57} \cdot  7^7 \cdot 57^{57} 
\]

\begin{table}[!h]
    \centering
    \caption{Concepts annotated in \BOIA. Table taken from \citep{xu2020boia}}
\begin{tabular}{llr}
\toprule
\textbf{Action Category}                 & \textbf{Concepts}             & \textbf{Count} \\ \hline
\multirow{3}{*}{$\texttt{move\_forward}$} & $\texttt{green\_light}$        & 7805           \\
                                         & $\texttt{follow}$              & 3489           \\
                                         & $\texttt{road\_clear}$         & 4838           \\ \hline
\multirow{6}{*}{$\texttt{stop}$}          & $\texttt{red\_light}$          & 5381           \\
                                         & $\texttt{traffic\_sign}$        & 1539           \\
                                         & $\texttt{car}$                 & 233            \\
                                         & $\texttt{person}$              & 163            \\
                                         & $\texttt{rider}$               & 5255           \\
                                         & $\texttt{other\_obstacle}$     & 455            \\ \hline
\multirow{6}{*}{$\texttt{turn\_left}$}          & $\texttt{left\_lane}$          & 154            \\
                                         & $\texttt{left\_green\_light}$  & 885            \\
                                         & $\texttt{left\_follow}$        & 365            \\ \cline{2-3}
                                         & $\texttt{no\_left\_lane}$      & 150            \\
                                         & $\texttt{left\_obstacle}$      & 666            \\
                                         & $\texttt{letf\_solid\_line}$   & 316            \\ \hline
\multirow{6}{*}{$\texttt{turn\_right}$}         & $\texttt{right\_lane}$         & 6081           \\
                                         & $\texttt{right\_green\_light}$ & 4022           \\
                                         & $\texttt{right\_follow}$       & 2161           \\ \cline{2-3}
                                         & $\texttt{no\_right\_lane}$     & 4503           \\
                                         & $\texttt{right\_obstacle}$     & 4514           \\
                                         & $\texttt{right\_solid\_line}$  & 3660  \\
\bottomrule
\end{tabular}
    \label{tab:boia-concepts}
\end{table}

\subsection{Optimizer and Hyper-parameter Selection}

The Adam optimizer \citep{KingmaB14@adam} was employed for all experiments, with exponential decay of the learning rate amounting to $\gamma=0.95$, exception made for \BOIA, where we added a weight-decay with parameter $\omega=4 \cdot 10^{-5}$ and $\gamma$ was set to $0.1$, to avoid over-fitting. 

The learning rate for all experiments was tuned by searching over the range $10^{-4}\div 10^{-2}$, with a total of $5$ log steps. We found the SL penalty of $2$ and $10$ for \XOR and \MNISTOp, respectively, to work well in our experiments. 
The strength of the single mitigation strategies ($\eta$) for each method was chosen accordingly to a grid-search over $\eta \in \{0.1,0.5,1,2,5,10\}$, varying the learning rate. The best hyper-parameters were selected, in the first step, based on the highest performances in $F_1$-score for label accuracy on the validation set and, in the second step, on the lowest mitigation loss, accordingly to the value in the validation set. 
In particular, for SL we chose those runs that yielded the best trade-off between SL minimization and label prediction. 
For LTN, we found that concept supervision did interfere with the original training objective, for which the best weight for the mitigation strength was found at $\gamma=10^{-2}$. It was adopted for both LTN+\MSC and LTN+\MSR+\MSC in \textbf{Q2} of \cref{tab:results-MNIST}.
The hyper-parameters for the combined mitigation strategies were selected according to the aforementioned criterion, by only searching the best-combined mitigation strength, while keeping the individual best strengths fixed from the previous grid search. 

For \BOIA, we selected the learning rate ranging in the interval $10^{-4}\div10^{-2}$ upon selecting those runs with best $F_1$-scores on labels in the validation set. The strength of the concept supervision and entropy regularization were varied in between $\{ 0.1, 1, 5 \}$. 

All best hyper-parameters for our tests are reported in the code in the Supplementary Material.

\subsection{Architectures} \label{sec:app-architectures}

\underline{\XOR}: For this data set, we adopted two MLPs, one for the encoder $p_\theta(\vC \mid \vG)$ and one for the decoder $p_\psi(\vG \mid \vC)$, both with a hidden size of $3$ and ReLU activations. \textit{This architecture is used to empirically validate RSs without forcing disentanglement.}
For the disentangled case, we considered a linear layer with weight $\omega$ and bias $b$.

For SL only, we added an additional MLP, with a hidden size of $3$ and $\tanh$ activations, implementing the map from the logits of $\vC$ to $\vY$.

\underline{\MNISTOp}: We report here the architectures that has been used for \MNISTAdd, \MNISTMul, and \MNISTShortcut. For the joint prediction, \ie without \textit{disentanglement}, we used the encoder in \cref{tab: mnist-addition-double-encoder}. 
When considering \textit{disentanglement}, we processed each digit with the encoder in \cref{tab: mnist-addition-single-encoder} and then stacked the two concepts together. For the reconstruction, we used the decoder in \cref{tab: mnist-addition-decoder}. 
For SL only, we added an MLP with a hidden size of $50$, taking as input the logits of both concepts and processing them to the label.

\begin{table}[!h]
    \centering
    \footnotesize
    \caption{Double digit encoder for \MNISTAdd}
    \begin{tabular}{cccc}
         \toprule
         \textsc{Input shape} & \textsc{Layer type} & \textsc{Parameters} & \textsc{Activation} \\ 
         \midrule
         $(28,56, 1)$ & Convolution & {depth}=$32$,  {kernel}=$4$, {stride}=$2$, {padding}=$1$ & ReLU \\
         $(32,14,28)$ & Dropout & $p=0.5$ & \\
         $(32,14,28)$ & Convolution & depth=$64$,  kernel=$4$, stride=$2$, padding=$1$ & ReLU \\
         $(64,7,14)$  & Dropout & $p=0.5$ & \\
         $(64,7,14)$  & Convolution & depth=$128$,  kernel=$4$, stride=$2$, padding=$1$ & ReLU \\
         $(128,3,7) $ & Flatten &   &  \\
         $(2688)$     & Linear & dim=$20$, {bias} = True \\
         \bottomrule
    \end{tabular}
    \label{tab: mnist-addition-double-encoder}
\end{table}

\begin{table}[!h]
    \centering
    \footnotesize
    \caption{Single digit Encoder for \MNISTOp}
    \begin{tabular}{cccc}
         \toprule
         \textsc{Input shape} & \textsc{Layer type} & \textsc{Parameters} & \textsc{Activation} \\ 
         \midrule
         $(28, 28, 1 )$ & Convolution & {depth}=$64$,  {kernel}=$4$, {stride}=$2$, {padding}=$1$ & ReLU \\
         $(14,14,64)$ & Dropout & $p=0.5$ & \\
         $(14,14, 64)$  & Convolution & depth=$128$,  kernel=$4$, stride=$2$, padding=$1$ & ReLU \\
         $(7,7,128)$ & Dropout & $p=0.5$ & \\
         $(7, 7, 128)$  & Convolution & depth=$256$,  kernel=$4$, stride=$2$, padding=$1$ & ReLU \\
         $(3,3,256) $ & Flatten &   &  \\
         $(2304)$ & Linear & dim=$10$, {bias} = True \\
         \bottomrule
    \end{tabular}
    \label{tab: mnist-addition-single-encoder}
\end{table}

\begin{table}[!h]
    \centering
    \footnotesize
    \caption{Decoder for \MNISTAdd}
    \begin{tabular}{cccc}
         \toprule
         \textsc{Input shape} & \textsc{Layer type} & \textsc{Parameters} & \textsc{Activation} \\ 
         \midrule
         $(40, 1)$            & Unflatten           &                     &  \\
         $(128,3,7)$          & ConvTranspose2d     & depth=$64$, kernel=$(5,4)$, stride=$2$, padding=$1$ & ReLU \\
         $(64,7,14)$          & Dropout & $p=0.5$ & \\
         $(128,3,7)$          & ConvTranspose2d     & depth=$32$, kernel=$(4,4)$, stride=$2$, padding=$1$ & ReLU \\   
         $(32,14,28) $        & Dropout & $p=0.5$ &  \\
         $(128,3,7)$          & ConvTranspose2d     & depth=$1$, kernel=$(4,4)$, stride=$2$, padding=$1$ & Sigmoid \\ 
         \bottomrule
    \end{tabular}
    \label{tab: mnist-addition-decoder}
\end{table}

\underline{\BOIA}: 
Images of \BOIA are preprocessed following \citep{sawada2022concept} with a Faster-RCNN \citep{ren2015faster} pre-trained on MS-COCO and fine-tuned on BDD-100k \cite{xu2020boia}. 
Successively, we adopted the pre-trained convolutional layer on \citep{sawada2022concept} to extract linear features, with dimension $2048$. These are the inputs for the NeSy model, which is implemented with a fully-connected NN, see \cref{tab:boia-backbone}.

\begin{table}[!h]
    \centering
    \footnotesize
    \caption{Fully connected layer for \BOIA}
    \begin{tabular}{cccc}
         \toprule
         \textsc{Input shape} & \textsc{Layer type} & \textsc{Parameters} & \textsc{Activation} \\ 
         \midrule
         $(2048, 1)$ & Linear + BatchNorm1d  & dim=1024 & Softplus \\
         $(1024)$ & Linear + BatchNorm1d  & dim=512 & Softplus \\
         $(512)$ & Linear + BatchNorm1d  & dim=256 & Softplus \\
         $(256)$ & Linear + BatchNorm1d  & dim=128 & Softplus \\
         $(128)$ & Linear + BatchNorm1d  & dim=21 & Softplus  \\
         \bottomrule
    \end{tabular}
    \label{tab:boia-backbone}
\end{table}

\clearpage
\newpage
\subsection{Confusion Matrices}
\label{sec:confusion-matrices}

Following, we report the label and concept-level confusion matrices (CMs) for \XOR, \MNISTShortcut, and \MNISTMul.  For all of them, we report those obtained in runs with maximal $F_1$-score over the labels.

\subsection{\XOR}

\begin{figure}[!h]
    \centering
    \begin{tabular}{cc}
        \textsc{none} & \DIS  \\
    \includegraphics[width=0.45\textwidth]{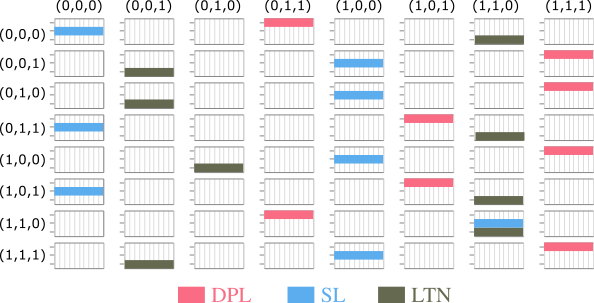} &
    \includegraphics[width=0.45\textwidth]{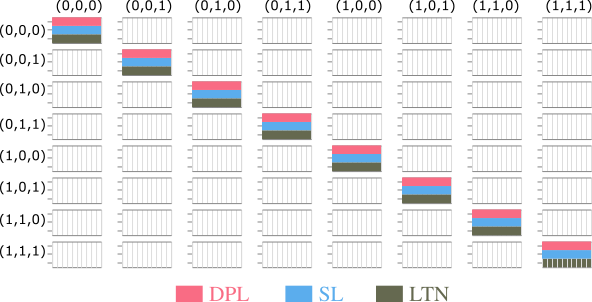}
    \end{tabular}
    \caption{\textbf{CMs for} \XOR: (\textit{Left}) All NeSy models fail for RSs without any mitigation. (\textit{Right}) Providing \DIS avoids all RSs.}
    \label{fig:XOR-CMs}
\end{figure}

\subsection{\MNISTShortcut}

\begin{figure}[!h]
    \centering
    \includegraphics[width=0.9\textwidth]{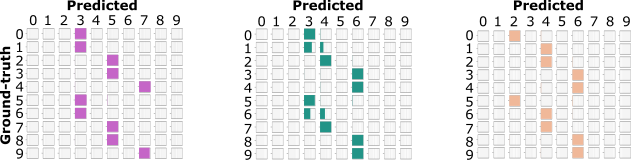}
    \caption{\textbf{NeSy models without mitigation strategies.} (\textit{Left}) DPL picks a RS that uses only $3$ digits. (\textit{Middle}) SL optimizes for label predictions but does not always predict a correct configuration for the digits. (\textit{Right}) LTN also picks a RS using only $3$ digits.}
\end{figure}
\begin{figure}[!h]
    \centering
    \includegraphics[width=0.9\textwidth]{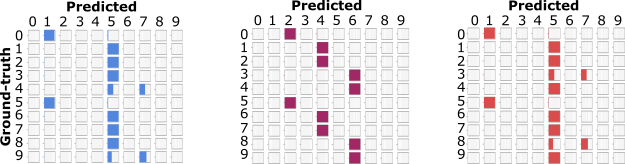}
    \caption{\textbf{NeSy models with \MSR.} (\textit{Left}) DPL picks a sub-optimal RS. (\textit{Middle}) SL optimizes for label predictions but through a RS. (\textit{Right}) LTN also picks a sub-optimal RS. For all runs, we found that \MSR interferes with the standard learning objective of DPL and LTN, respectively.}
\end{figure}
\begin{figure}[!h]
    \centering
    \includegraphics[width=0.9\textwidth]{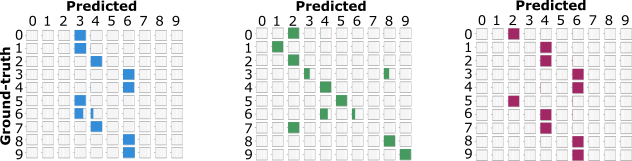}
    \caption{\textbf{NeSy models with \MSC.} (\textit{Left}) DPL picks a sub-optimal RS and fails to predict correctly the digits $4$ and $9$. (\textit{Middle}) SL predicts correctly the $4$'s and $9$'s but does not avoid the RS. (\textit{Right}) LTN picks a RS that uses only $3$ digits.}
\end{figure}
\begin{figure}[!h]
    \centering
    \includegraphics[width=0.9\textwidth]{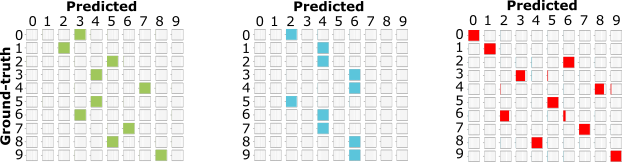}
    \caption{\textbf{NeSy models with \MSH.} (\textit{Left}) DPL picks a RS that uses concepts more sparsely. (\textit{Middle}) SL picks a RS, irrespectively of the mitigation. (\textit{Right}) LTN tends to align to the diagonal but fails to predict correctly multiple digits. The performance, nonetheless, is sub-optimal.}
\end{figure}
\begin{figure}[!h]
    \centering
    \includegraphics[width=0.9\textwidth]{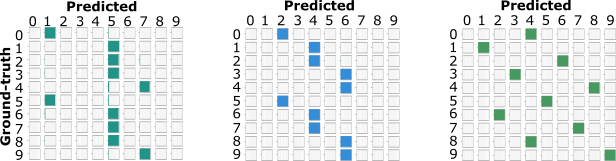}
    \caption{\textbf{NeSy models with \MSR and \MSH.} (\textit{Left}) DPL picks a sub-optimal RS by using only three digits. (\textit{Middle}) SL picks a RS. (\textit{Right}) LTN learns correctly the odd digits but learns a RS for the even ones.}
\end{figure}
\begin{figure}[!h]
    \centering
    \includegraphics[width=0.9\textwidth]{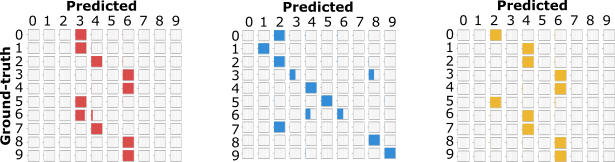}
    \caption{\textbf{NeSy models with \MSR and \MSC.} (\textit{Left}) DPL picks a sub-optimal RS by using only three digits. (\textit{Middle}) SL improves along the diagonal but fails to correctly encode four digits. (\textit{Right}) LTN picks a RS.}
\end{figure}
\begin{figure}[!h]
    \centering
    \includegraphics[width=0.9\textwidth]{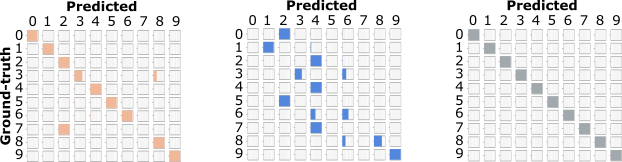}
    \caption{\textbf{NeSy models with \MSC and \MSH.} (\textit{Left}) DPL correctly retrieves almost all digits but fails with the digit $7$. For this method, we also found runs completely recovering the diagonal. (\textit{Middle}) SL picks a sub-optimal RS. (\textit{Right}) LTN avoids the RS. }
\end{figure}
\begin{figure}[!h]
    \centering
    \includegraphics[width=0.9\textwidth]{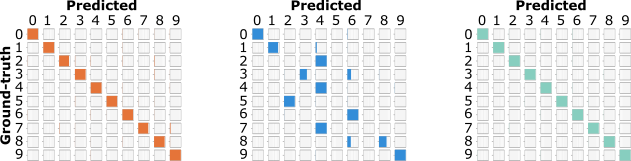}
    \caption{\textbf{NeSy models with \MSR, \MSC and \MSH.} (\textit{Left}) DPLcorrectly identifies the underlying concepts. (\textit{Middle}) SL does not avoid RSs. (\textit{Right}) LTN also identifies the correct digits.}
\end{figure} 

\clearpage
\newpage
\subsection{\MNISTOp}

\begin{figure}[!h]
    \centering
    \includegraphics[width=0.8\textwidth]{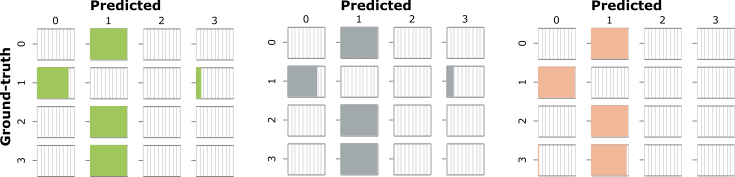}
    \caption{\textbf{CM on the addition task.} All models, DPL (\textit{Left}), SL (\textit{Middle}), and LTN (\textit{Right}) fail for the RS introduced in \cref{ex:mnist-addition}, while failing also to predict the digit $3$. }
\end{figure}

\begin{figure}[!h]
    \centering
    \includegraphics[width=0.8\textwidth]{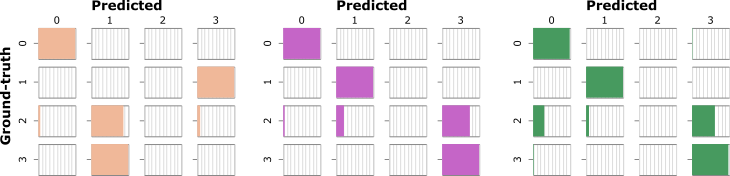}
    \caption{\textbf{CM on the product task.} (\textit{Left}) DPL picks a RS where it fails to capture the correct semantics of all digits, except $0$, (\textit{Middle}, \textit{Right}) SL, and likewise LTN, acquires a RS where it never predicts correctly the digit $2$.}
\end{figure}

\begin{figure}[!h]
    \centering
    \includegraphics[width=0.8\textwidth]{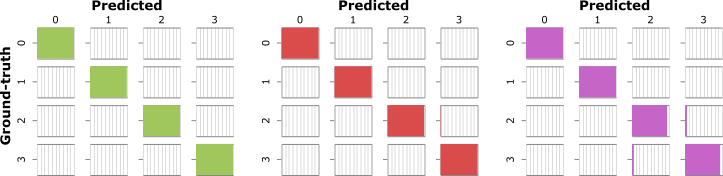}
    \caption{\textbf{CM on \MTL.} Once the predictors are trained for solving addition and multiplication \textit{jointly} through \MTL, they all successfully acquire the concepts with the intended semantics: all confusion matrices are very close to being diagonal.}
\end{figure}

\newpage

\end{document}